\documentclass[preprint,12pt]{article}
\usepackage{geometry}
\usepackage{amssymb}
\usepackage{bm}
\usepackage{algorithm}
\usepackage{algorithmic}
\usepackage{amsthm}
\usepackage{amssymb}
\usepackage{amsmath}
\usepackage{graphicx}
\usepackage{subfigure}
\usepackage{booktabs}
\usepackage{hyperref}

\newcommand{\subheading}[1]{
  \vspace{0.5cm}
  \noindent {\bf #1.}
}

\newcommand{\hz}{y}

\newcommand{\ve}[1]{\mathbf{#1}}
\newcommand{\W}{\mathcal{W}}

\newcommand{\ddthetai}{\frac{\partial}{\partial \theta_{i}}}

\newcommand{\dd}[1]{\frac{\partial}{\partial #1}}

\newcommand{\bGamma}{\boldsymbol{\Gamma}}
\newcommand{\bth}{\boldsymbol{\theta}}
\newcommand{\thi}{\theta_i(t)}

\newcommand{\pn}[2]{p_{\mathcal{N}}\left({#1}\,\middle\vert\,{#2}\right)}

\newcommand{\ps}[1]{p_{\mathcal{S}}\left({#1}\right)}

\newcommand{\zk}{z_{\posti}}
\newcommand{\zkt}{\zk(t)}
\newcommand{\zks}{\zk(s)}

\newcommand{\fk}{f_{\posti}}

\newcommand{\fks}{\fk(s)}

\newcommand{\bbz}{\mathbf{z}}

\newcommand{\bbzt}{\mathbf{z}(t)}
\newcommand{\bbxt}{\mathbf{x}(t)}

\newcommand{\bbztau}{\mathbf{z}(\tau)}

\newcommand{\pl}[1]{{\bf #1:}}
\newcommand{\pt}{p^*_T(\bth)}
\newcommand{\camk}{CaMKII}

\newcommand{\choosealt}[4]{
\left\{
  \begin{array}{lll}
    #1 & , & \mbox{ for } #2\\
    #3 & , & \mbox{ for } #4\\
  \end{array}
\right.}

\usepackage[]{caption2}

\newcommand{\syn}[1]{\text{\textsc{syn}}_{#1}}
\newcommand{\prei}{\text{\tiny{\textsc{pre}}}_{i}}
\newcommand{\posti}{\text{\tiny{\textsc{post}}}_{i}}
\newcommand{\preidef}{\text{\textsc{pre}}_{i}}

\renewcommand{\d}{d}

\newcommand{\expect}[2][]{\left\langle\, #2\, \right\rangle_{#1} }



\newcommand{\cprob}[2]{p\left({#1}\,\middle\vert\,{#2}\right)}
\newcommand{\wiener}{\mathcal{W}}

\usepackage{setspace}%

\title{\camk{} activation supports reward-based neural network optimization through Hamiltonian sampling}

\author{
Zhaofei Yu$^{1,*}$, David Kappel$^{2,*}$, Robert Legenstein$^{2,*}$,  Sen Song$^{3}$, \\
Feng Chen$^{1}$  and Wolfgang Maass$^{2} $ \\ \\
\begin{minipage}{8cm}
\small \centering
1) \quad  Department of Automation\\
Tsinghua University\\
100084 Beijing, China\\
\texttt{chenfeng@mails.tsinghua.edu.cn} \\
\end{minipage}
\begin{minipage}{8cm}
\small \centering
2) \quad Institute for Theoretical Computer Science\\
Graz University of Technology\\
8010 Graz, Austria \\
\texttt{maass@igi.tugraz.at} \\
\end{minipage}\\
\quad\\
\begin{minipage}{8cm}
\small \centering
\small 3) \quad Department of Biomedical Engineering\\
\small  Tsinghua University\\
\small  100084 Beijing, China\\
\small  \texttt{sen.song@gmail.com} \\
\end{minipage} \\
\quad\\
\small *) \quad These authors contributed equally to this work
}

\begin{document}
\maketitle

\begin{abstract}
Synaptic plasticity is implemented and controlled through over thousand different types of molecules in the postsynaptic density and presynaptic boutons that assume a staggering array of different states through phosporylation and other mechanisms. One of the most prominent molecule in the postsynaptic density is \camk{}, that is described in molecular biology as a ``memory molecule'' that can integrate through auto-phosporylation Ca-influx signals on a relatively large time scale of dozens of seconds. The functional impact of this memory mechanism is largely unknown. We show that the experimental data on the specific role of \camk{} activation in dopamine-gated spine consolidation suggest a general functional role in speeding up reward-guided search for network configurations that maximize reward expectation. Our theoretical analysis shows that stochastic search could in principle even attain optimal network configurations by emulating one of the most well-known nonlinear optimization methods, simulated annealing. But this optimization is usually impeded by slowness of stochastic search at a given temperature.  We propose that \camk{}  contributes a momentum term that substantially speeds up this search. In particular, it allows the network to overcome saddle points of the fitness function. The resulting improved stochastic policy search can be understood on a more abstract level as Hamiltonian sampling, which is known to be one of the most efficient stochastic search methods. 
\end{abstract}

\section{Introduction}

\begin{figure}
\begin{center}
\includegraphics{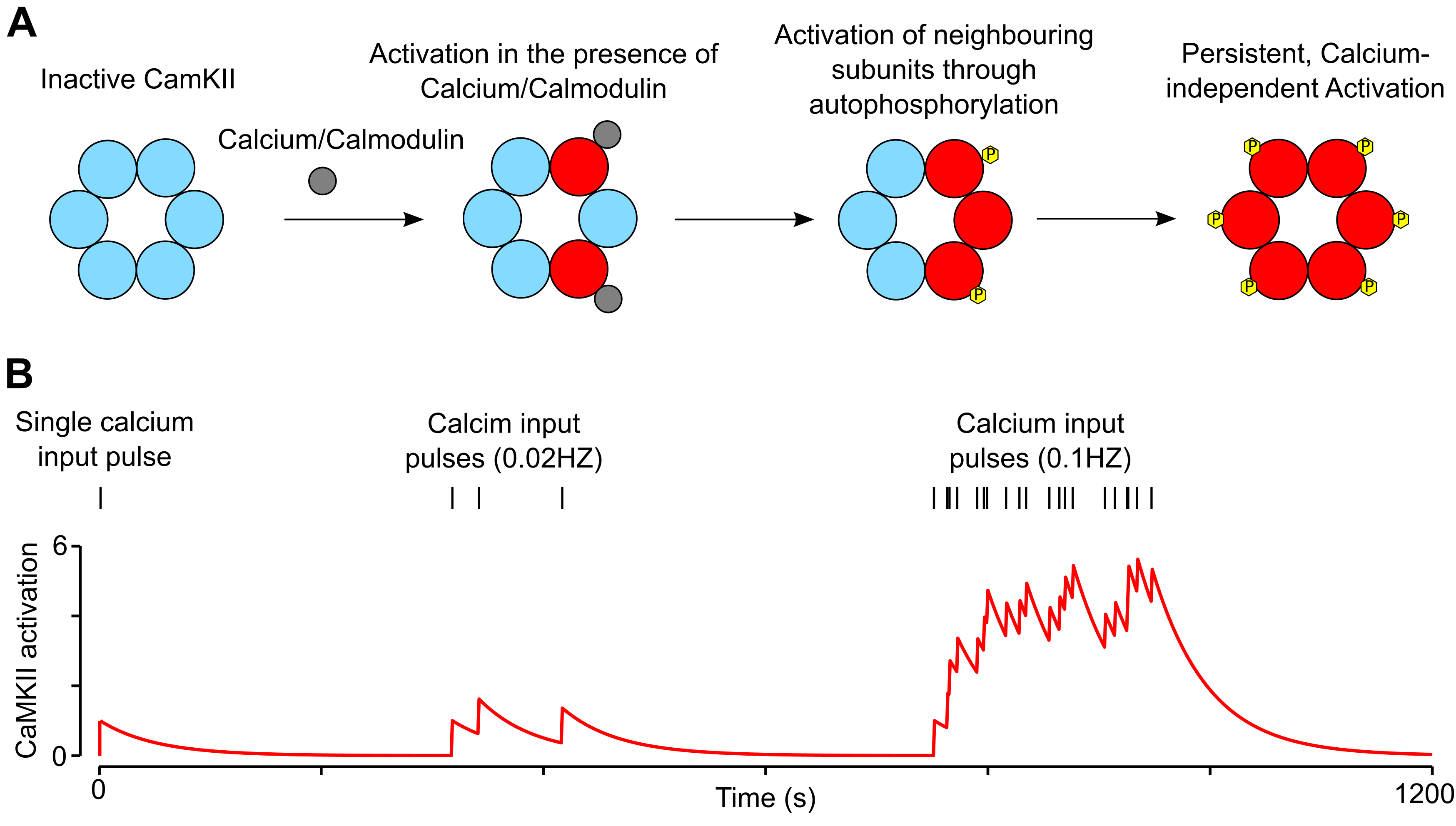}
\end{center}
\caption{{\bf {\camk{}} dynamics.}
\pl{A} \camk{} has a crystal structure of 12 units, which fold into two rings of six domains. Incoming calcium/calmodulin (grey) can transform \camk{} into its active state (red), and then activate the neighboring subunits through autophosphorylation. This leads to persistent and calcium-independent activation of the \camk{} on the time scale of dozens of seconds (figure adapted from \cite{Mhbrugman2011}).
\pl{B} The local concentration of \camk{} in its activated state changes in response to different input pulse. \camk{} activation jumps upward by 1 with a single calcium input pulse and otherwise decays exponentially. For Poisson calcium input pulses, \camk{} activation is irregular for low frequency input (0.02HZ) and almost steady for high frequency input (0.1HZ). Note that the time constant of \camk{} is set to 50 s here.
} 
\label{fig:camk}
\end{figure}

Calcium-calmodulin dependent protein kinase II (\camk{}) is the most frequently occurring complex molecule in the postsynaptic density and a key molecule for the implementation of synaptic plasticity \cite{sheng2011postsynaptic} (see Fig.~\ref{fig:camk}A). It is described in molecular biology as a ``memory molecule'' that creates through its somewhat persistent autophosphorylated (active) state a short term memory or low pass filter on the time scale of dozens of seconds for calcium influx (see e.g.~Ch.~15 in \cite{albert2014molecular}, Fig.~1c in \cite{lisman2012mechanisms}, and Fig.~3F in \cite{YagishitaETAL:14}). Calcium influx is a typical feature of the induction of longterm plasticity via NMDA receptors. More specifically, incoming calcium transforms \camk{} via calmodulin into its active state, which is maintained for a while via autophosphorylation among its 12 subunits. Furthermore \camk{} triggers in its activated state changes of synaptic efficacy through the phosphorylation of AMPA receptors, the anchoring of additional AMPA receptors in the postsynaptic density, and dopamine-gated stabilization of spines (see e.g.~Fig.~3, S5, S11 in \cite{YagishitaETAL:14}). 

Although numerous experimental data show that \camk{} in its activated state is essential both for LTP and LTD \cite{coultrap2014autonomous,connor_wang_2015}, its contribution to network plasticity is still unclear.
We address in this article the question how the activation dynamics of \camk{} could contribute to reward-based network optimization for specific computational tasks.
Since the molecular processes that involve \camk{} and give rise to LTP and LTD contain a strong stochastic component, it is natural to view this optimization not as a deterministic but a stochastic search for good network parameters. 
This view is also consistent with numerous experimental data that show that synaptic connections are even in the adult cortex subject to a continuous coming and going of dendritic spines that appears to be inherently stochastic and independent of pre- or postsynaptic firing in the absence of a functional synaptic connection \cite{yasumatsu2008principles,holtmaat2009experience}. 
A theoretical framework for stochastic network plasticity has been introduced in \cite{kappel2015network,kappel2016reward} and termed synaptic sampling. There, it was shown that a neural network $\mathcal{N}$ with parameters $\bth$ subject to stochastic plasticity rules samples from a stationary distribution $p_T^*(\bth)$ of network configurations through a sampling process known as Langevin sampling in the machine learning literature. This means that the network will visit -- in the long run -- network configurations $\bth$ most often that have a large probability $p_T^*(\bth)$. The index $T$ in the distribution $p_T^*(\bth)$ denotes the ``temperature'' of the search, which depends on the amount of noise in the plasticity process. The exact shape of $p_T^*(\bth)$ is determined by the plasticity rule and in the context of reward-based learning it can be chosen to prefer network configurations that lead frequently to large rewards. In other words, in this framework, synaptic plasticity can be shown to implement an ongoing stochastic policy search \cite{kappel2016reward}. 

However, as synaptic sampling carries out Langevin sampling, convergence to the stationary distribution is rather slow for any fixed temperature, which is in general undesirable as it implies slow learning. In particular, the search for high fitness regions by gradient-based optimization techniques such as Langevin sampling is hindered by local optima and -- even more severely as recently suggested by Dauphin et al.~\cite{dauphin2014identifying,dauphin2015equilibrated} -- by the presence of saddle points in $p_T^*(\bth)$.
This slowness is a generic impediment for the implementation of a global optimization strategy such as simulated annealing as further detailed below. 

In this article, we show that \camk{} activation dynamics can ease these issues. Compared to the synaptic sampling framework, the activation dynamics of \camk{} gives rise to an additional dynamic variable that basically low-pass filters parameter updates. This low-pass filtering implements a momentum term, a method that is well-known to improve gradient-based optimization in many circumstances, for example in the vicinity of saddle points. More abstractly, in our stochastic framework,
we show that the resulting dynamics gives rise to a parameter sampling algorithm known as Hamiltonian sampling, that however still samples from the same stationary distribution $p_T^*(\bth)$. A well-known advantage of Hamiltonian sampling over Langevin sampling is faster convergence to the stationary distribution \cite{sutskever2013importance}.

With such faster convergence properties, our model for \camk{} driven plasticity allows us to create a link from reward-based learning to optimization theory, which establishes conditions under which a neural circuit could attain not only functionally attractive locally optimal network configurations, but in principle even a global optimum. Simulated annealing \cite{aarts1988simulated,dekkers1991global} is arguably one of the most powerful algorithmic approach to nonlinear optimization. Evolutionary algorithms also work well in some cases, but require a large control overhead and many competing networks in parallel for which no biological evidence exists so far. We show that reward-based network plasticity can in principle reach even globally optimal network configurations $\bth$ if the amount of stochasticity is sufficiently slowly decreased during learning (``cooling'' or ``annealing''), similar to simulated annealing in continuous time. This theoretical result provides a new gold standard for reward-based network learning.

\section{Results}
\label{sec:results}
Consider a network $\mathcal{N}$ that receives at certain times $t$ reward signals $r(t)$, e.g., in the form of dopamine. 
The dynamics of each synaptic connection $i$ in the network $\mathcal{N}$ is modeled by a parameter $\theta_i(t)$, which determines the synaptic efficacy. Therefore, we assume that the behavior of the network (i.e., its response to network input; also referred to as the network policy) is determined by the parameter vector $\bth$ (the vector of all synaptic parameters). 
In biological neuronal networks, neurons are either excitatory or inhibitory, a fact that is commonly refered to as Dale's principle. This implies that their outgoing synaptic connections are exclusively excitatory (modelled as positive synaptic weights) or inhibitory (negative synaptic weights), and that these synaptic weights cannot change the their sign through plasticity processes.
We will first introduce a version of the model that allows such a sign-switch of synaptic weights for demonstration purposes. We will later introduce a slightly modified version of the model where only excitatory synapses are plastic with weights constrained to be non-negative.

 Previous work \cite{kappel2016reward} has
analyzed under which conditions such a network can perform an ongoing stochastic policy search. That is, under which conditions local stochastic synaptic plasticity processes on $\bth$ can achieve that
the network $\mathcal{N}$ seeks network configurations 
that provide a large expected discounted reward. Mathematically, the expected discounted reward $\mathcal{V}(\bth)$ for a given parameter vector $\bth$ is given by
\begin{equation}
  \mathcal{V}(\bth) \;=\; \expect[p(\ve r | \bth)]{ \int_{0}^\infty e^{-\frac{\tau}{\tau_e}} \,r(\tau) \; \d \tau } \;.
  \label{eqn:reward-prob-factorized}
\end{equation}
The integral integrates all future rewards $r(\tau)$, while discounting more remote rewards exponentially with a discount rate $\tau_e$.
The expectation is an average over multiple learning episodes where in each episode one realization of the reward trajectory $\ve r$ is encountered for the given parameters $\bth$ according to some distribution $p(\ve r | \bth)$.

In addition, a biological network $\mathcal{N}$ needs to satisfy structural constraints, such as sparse connectivity, that can be formulated through a prior $\ps{\bth}$ over network configurations $\bth$ \cite{kappel2016reward}. Hence, network learning can be regarded as a search for policies (i.e.,~network configurations $\bth$) that both satisfy structural constraints and provide a large expected discounted reward. This can be stated more formally as a sampling from the posterior distribution $p^*(\bth)$ of parameters
\begin{equation}
  p^*(\bth) \propto  \ps{\bth} \, \mathcal{V}(\bth).
  \label{eq:posterior}
\end{equation}
It was shown in \cite{kappel2016reward} that if the stochastic dynamics of each parameter $\theta_i$ can be characterized through a stochastic differential equation (SDE) of the form
\begin{align}
d \theta_i \;=\; \beta \, \left( \ddthetai \,\log p^*(\bth) \right)  dt  \;+ \; \sqrt{2 T \beta} \, d \wiener_{i} \;,
\label{eq:sde_rmsynsam}
\end{align}
\noindent then the network reaches the unique stationary distribution given by the posterior $p_T^*(\bth) = \frac{1}{{\cal Z}}p^*(\bth)^\frac{1}{T}$ and then samples from this distribution over network configurations.
The parameter $\beta>0$ denotes a learning rate that controls the speed of the parameter dynamics.
The last term $d \wiener_{i}$ of Eq.~\eqref{eq:sde} describes infinitesimal stochastic increments and decrements of a Wiener process $\wiener_{i}$ -- a standard model for Brownian motion in one dimension (see \cite{gardiner2004handbook}). The amplitude of this noise term is scaled by the temperature parameter $T>0$, which can be used to increase or decrease random exploration of the parameter space.

\begin{figure}
\begin{center}
\includegraphics[scale=0.95]{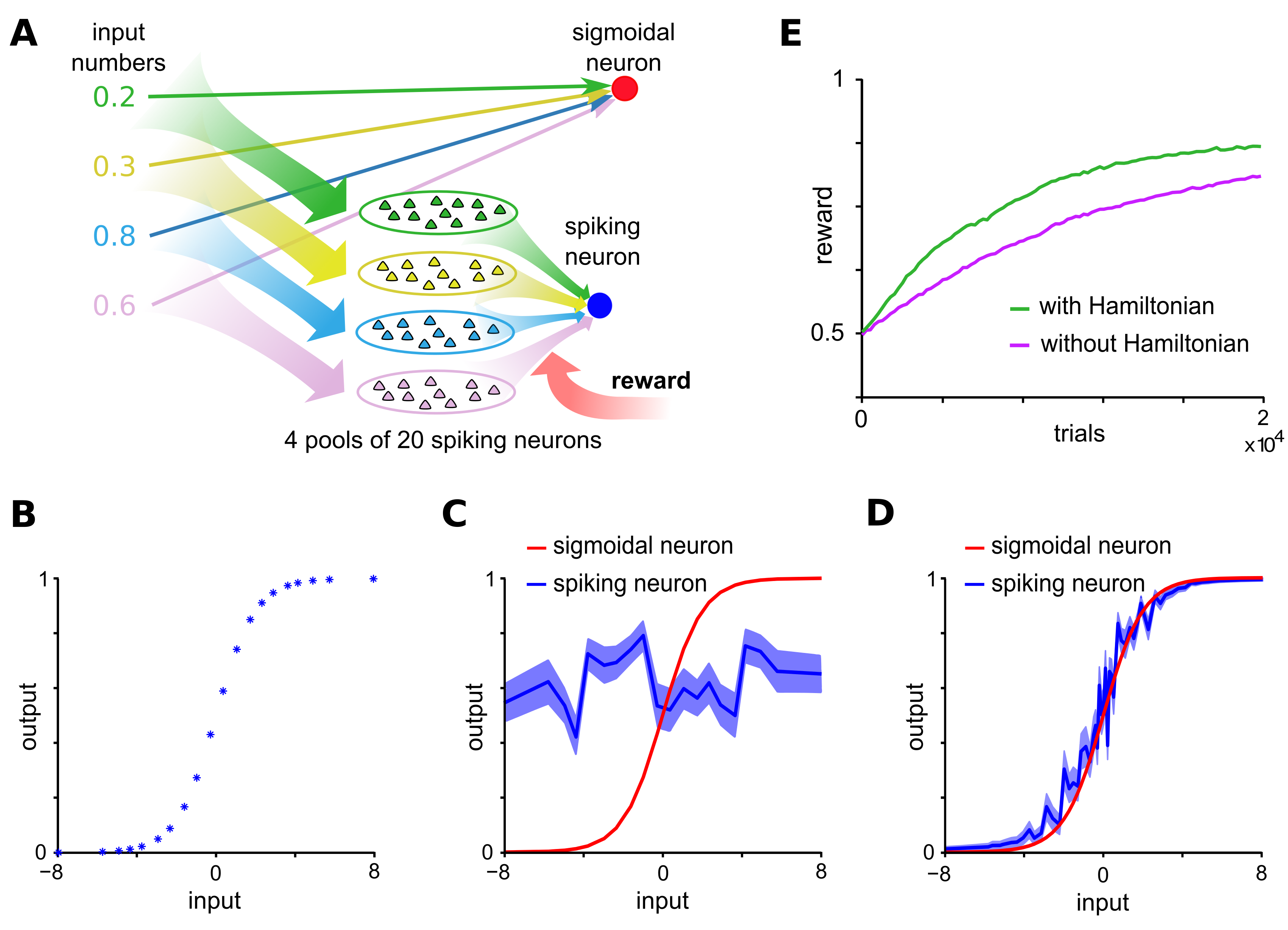}
\end{center}
\caption{{\bf A spiking neuron learns to emulate a sigmoidal neuron.}
\pl{A} Illustration of the network architecture. The target firing activity of the spiking neuron (blue) is defined by the output of a sigmoidal neuron (red) with four inputs and pre-defined weights. The spiking neuron receives inputs from 4 pools of 20 spiking neurons each, with firing rates proportional to the sigmoidal neurons' inputs. 
\pl{B} The distribution of the 20 input pattern used during learning on the input-output plane of the sigmoidal neuron (x-axis:  weighted sum of the four inputs).
\textbf{C}, \pl{D} Output of sigmoidal neuron (red) and firing probability of spiking neuron (blue) as a function of the weighted sum of inputs before (C) and after (D) learning through Hamiltonian dynamics. The spiking neuron approximates the smooth behavior of the sigmoidal neuron after learning.
\pl{E} Comparison of the average rewards for synaptic sampling with (green) and without (magenta) Hamiltonian
dynamics throughout learning (average over 50 trials).
} 
\label{fig:sigmoid}
\end{figure}

To integrate the role of \camk{} in the plasticity processes, we model the previously sketched transient role of \camk{} as a low pass filter in the induction of synaptic plasticity (see Fig.~\ref{fig:camk}B). For each potential synapse $i$, we introduce another dynamic variable $\Gamma_i(t)$ that determines the change of the $\thi$ at time $t$.
It was found that both, LTP and LTD require the activated form of \camk{}, and that the switch between LTP and LTD is determined by other mechanisms \cite{coultrap2014autonomous,connor_wang_2015}.  We therefore interpret the absolute value of $\Gamma_i(t)$ as the local concentration of \camk{} in its activated state.
The interaction of these two variables is modeled by the stochastic differential equation (SDE) of the form
\begin{equation}
\begin{array}{l}
d \theta_i \;=\;a~\Gamma_i~dt\\
d\Gamma_i\;=\;  \, \left( a\, \ddthetai  \,\log p^*(\bth)  \, -b\Gamma_i  \right)  dt  \;+ \; \sqrt{2 T b} \,d\W_{\Gamma_{i}}~ \;,
\label{eq:sde}
\end{array}
\end{equation}
where the change of parameter $\theta_i$ directly depends on the value of the hidden CaMKII-related variable $\Gamma_i$ ($a>0$ is a learning rate). The dynamics of $\Gamma_i$ in turn is determined by three terms. The first term is the gradient of the parameter posterior. In reward-based learning, this gradient can be estimated by a rule that depends only on pre- and post-synaptic spike times and a global reward signal implemented for example as a dopaminergic signal. The friction term $-b\Gamma_i$ implements the decay of \camk{} activation with a time constant $b$. Detailed experimental studies suggest that this time constant depends on a variety of factors, e.g. the inactivation time constant of \camk{} activity and the mobility of \camk{} \cite{lee2009activation,li2012calcium,bhattacharyya2016molecular} (we used 10~s in Fig.~\ref{fig:sigmoid} and 50~s in the remainder of the paper). The last term models noise on \camk{} activation, such as stochastic opening of N-methyl-d-aspartate (NMDA) receptor channels \cite{zeng2010effect}.

With these extended parameter dynamics, the network samples from the posterior $p^*_T(\bth, \bGamma) =\frac{1}{{\cal Z}}p^*{\left( {\bth } \right)^{\frac{1}{T}}}p^*{\left( {\bGamma } \right)^{\frac{1}{T}}}$ over network configurations (see Theorem 1 in \nameref{sec:methods} for details). By marginalization over the \camk{} parameters $\bGamma$ it then follows that the stationary distribution over the synaptic parameters again is given by $\pt = \int p^*_T(\bth, \bGamma)~d\bGamma = \frac{1}{\cal Z} p^*{\left( \bth  \right)^{\frac{1}{T}}}$.

In other words, the CaMKII-enriched dynamics gives rise to the same reward-optimizing distribution over network configurations as the direct dynamics considered in \cite{kappel2016reward}. Importantly however, it turns out that the dynamics \eqref{eq:sde} actually posesses advantageous properties when compared to the direct dynamics \eqref{eq:sde_rmsynsam}. 
For the noise-less case ($T=0$), the dynamics \eqref{eq:sde_rmsynsam} corresponds to a gradient ascent on $p^*(\bth)$. In comparison, the dynamics \eqref{eq:sde} introduces a momentum term which is well-known to improve gradient descent in many circumstances, for example in the presence of small local optima or in the vicinity of saddle points. In the case with noise, the dynamics \eqref{eq:sde_rmsynsam} corresponds to Langevin sampling from $p^*(\bth)$, and the dynamics \eqref{eq:sde} to Hamiltonian sampling with friction. It is knwon that Hamiltonian sampling typically shows much faster convergence to the stationary distribution than the rather slow Langevin sampling \cite{neal2011mcmc}. In fact, a similar low-pass filtering of gradient updates was already implemented in \cite{kappel2016reward} to improve learning performance, but without a clean mathematical background and biological motivation.

To arrive at concrete plasticity rules, one has to determine $\ddthetai  \,\log p^*(\bth) = \ddthetai \log \ps{\bth} + \ddthetai \log \mathcal{V}(\bth)$ in Eq.~\eqref{eq:sde}, for the concrete neuron model and prior $\ps{\bth}$ at hand. As one example to be used in subsequent simulations, we consider a stochastic spiking neuron model (see \nameref{sec:spiking-neuron-model}) and independent zero-mean Gaussian priors with variance $\sigma^2$ for each parameter $\theta_{i}$. We obtain $\ddthetai \log \ps{\bth} = - \frac{1}{\sigma^2} \theta_{i}$ for the derivative of the prior. Using this and Eq.~\eqref{eqn:reward-prob-factorized} we find that the derivative $\ddthetai  \,\log p^*(\bth)$ in Eq.~\eqref{eq:sde} at time $t$ can be approximated by (see \nameref{sec:methods})
\begin{eqnarray}
  & \ddthetai  \,\log p^*(\bth) \;=\; \ddthetai \log \ps{\bth} + \ddthetai \log \mathcal{V}(\bth) \;\approx\; r(t) \, e_{i}(t) - \frac{1}{\sigma^2} \theta_{i}(t)  \label{eqn:gradient-est} \\
  & \frac{de_{i}(t)}{dt} = -\frac{1}{\tau_e}\,~e_{i}(t) \;+\; \hz_{\prei}(t) \,  (\zkt -f_{\posti}(t)) \; , \label{eqn:eligibility-trace} 
\end{eqnarray}
where $\hz_{\prei}(t)$ is the PSP activation under synapse $i$, $f_{\posti}(t)$ denotes the firing probability of the postsynaptic neuron, and $\zkt$ is a binary variable that is one if the postsynaptic neuron spiked at time $t$ and zero else.
Here the synaptic plasticity rule acts on $\Gamma_i$ (see Eq.~\eqref{eq:sde}) which is related to CaMKII activation instead of acting directly on the synaptic parameter $\theta_i$.
This learning rule is a simple version of reward-modulated spike-timing dependent plasticity (STDP). Similar rules were derived previously in the context of reward-based learning \cite{urbanczik2009reinforcement, brea2013matching}. The current work extends these rules to include a prior over network configurations, stochastic parameter updates, and CaMKII-induced Hamiltonian dynamics.

\subsection{A spiking neuron learns to emulate a sigmoidal neuron}

Learning in recurrent networks of spiking neurons is notoriously hard \cite{abbott2016building}, in particular with reward-based learning. For example, interesting functionality has been acquired through reward-based synaptic plasticity in \cite{hoerzer2014emergence}, but only in recurrent networks of smooth non-spiking neurons. Recently, it has been proposed that functionality of a non-spiking network can be ported to a spiking network if it has previously learned to exhibit smooth dynamics \cite{abbott2016building}. We wondered whether such smoothing of network responses can be obtained through reward-based learning. We considered a very simple basic setup where the task is to reproduce with a single spiking neuron the behavior of an artificial sigmoidal neuron model (see Fig.~\ref{fig:sigmoid}A).

The target firing rate of the spiking neuron was given by the output of a sigmoidal neuron with four inputs and pre-defined weights. Fig.~\ref{fig:sigmoid}B shows the desired input-output behavior. The spiking neuron received inputs from 4 pools of 20 spiking neurons each, with firing rates proportional to the sigmoidal neurons' inputs (and a maximum of $60$ Hz).
Input patterns were presented to the spiking neuron continuously while its weights were adapted through reward-based plasticity (at each presentation, one out of $20$ patterns was chosen randomly and presented, see Fig.~\ref{fig:sigmoid}B). Each presentation of an input pattern lasted for $300$ ms. The presentation was followed by a $10$ ms phase where a reward was delivered which was given by 1 minus the absolute difference between spiking neuron and sigmoidal neuron output (see \nameref{sec:methods} for details). After reward delivery, a $400$~ms delay period was introduced where input neurons were silent, followed by another pattern presentation.

Before learning, the firing rate of the spiking neuron was rather random over the whole range of inputs (Fig.~\ref{fig:sigmoid}C). After 20000 pattern presentations, the neurons' firing rate approximated the smooth behavior of the sigmoidal neuron well (Fig.~\ref{fig:sigmoid}D). Fig.~\ref{fig:sigmoid}E shows the average reward throughout learning for Hamiltonian sampling in comparison with non-Hamiltionian dynamics (synaptic sampling). One can see that Hamiltonian dynamics speeds up learning significantly.

\subsection{Reward-guided network plasticity}
\label{sec:peters}

\begin{figure}
\begin{center}
\includegraphics[scale=1]{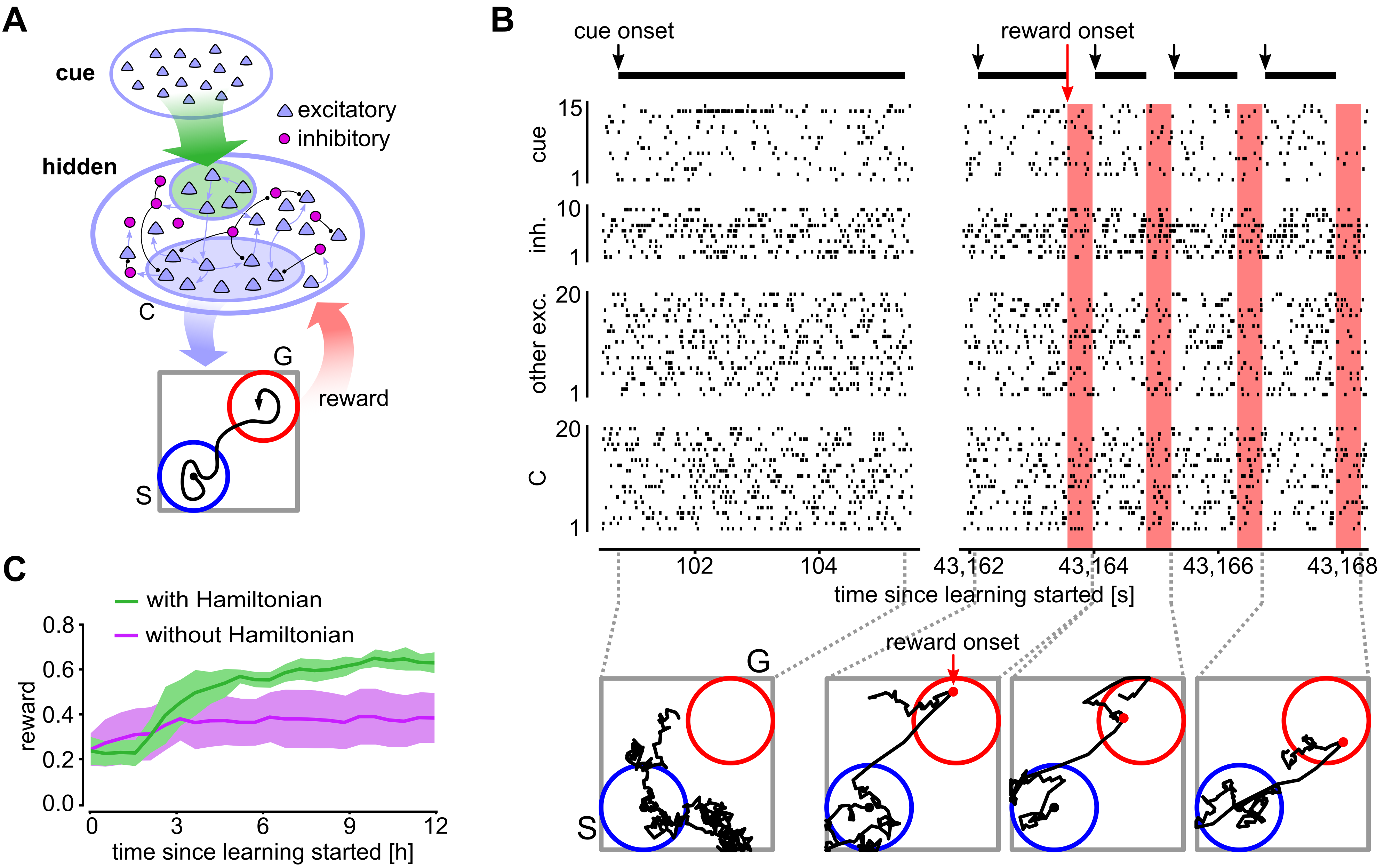}
\end{center}
\caption{{\bf {Hamiltonian synapse dynamics improves learning a blind reaching task.}}
\pl{A} Illustration of the network architecture and learning task. A recurrent network of inhibitory and excitatory neurons with input from a population of afferent neurons. The arrows indicate symbolically the connectivity between the excitatory neurons (blue), and inhibitory neurons (purple) (random subsets are shown). A pool C of neurons is used to control the position of the cursor in 2D space. The afferent input neurons provide a cue that indicates the phase during which the movement should be performed. Reward is delivered to the network if the cursor reaches the goal location G starting from the start location S.
\pl{B} Activity from random subsets of the network neurons (top) and example cursor trajectories (bottom) at learning onset time and after 12 hours of learning. The black horizontal bars indicate the presentation of the cue pattern. The red vertical bars show the reward windows at the end of successful trials. Network responses and cursor movements become more stereotyped and goal-directed throughout learning.
\pl{C} Comparison of learning curves with (green) and without (magenta) Hamiltonian synaptic dynamics. Reward is quantified here by the mean fraction of successful trials at each time point. If Hamiltonian dynamics is included the network learns the task significantly faster and better.  Average results over 5 independent trials are shown, shaded area indicates STD.
} 
\label{fig:Peter}
\end{figure}

Next we investigated whether the benefit in learning performance of Hamiltonian sampling scales up to biologically more realistic network architectures, that are larger in size and less structured. To do so, we applied the Hamiltonian synaptic sampling framework outlined above to learn a blind reaching task in a simple model of motor cortex. Reward-guided changes of network activity and task-induced spine dynamics are well documented in motor cortex \cite{PetersETAL:14}. We used a network of 100 recurrently connected excitatory neurons and 20 inhibitory neurons to control a cursor in 2D space  (see Fig.~\ref{fig:Peter}A,B). Connectivity parameters of this cortical network motif were taken from \cite{AvermannETAL:12} (see \nameref{sec:methods}). In addition to recurrent connections a random subset of 30 excitatory neurons received input from 200 afferent neurons. From the remaining 70 excitatory neurons we randomly selected a neural pool C of 50 neurons to control the cursor position. For controlling the cursor we adopted the population vector model \cite{georgopoulos1986neuronal}. Briefly, each neuron in C was assigned a randomly selected preferred direction in 2D cursor space. At each time point the cursor was moved in the direction of the population vector (accumulated preferred directions weighted by neural activities) of the 50 neurons in C.

Each trial started with the cursor centered at the start area S (blue circle in Fig.~\ref{fig:Peter}A). The cursor had to be held at S for 50~ms to initiate the movement phase of the trial. The movement phase was indicated through the presentation of a cue pattern (a rate pattern for all 200 afferent input neurons, see \nameref{sec:methods}). Reward was given to the network if the cursor was moved to the target area G in Fig.~\ref{fig:Peter}A and held there for 50~ms. At success, the presentation of the cue pattern was stopped and a 400~ms reward window was initiated during which $r(t)$ was set to $1$ (indicated by red vertical bars in Fig.~\ref{fig:Peter}B). If the network failed to reach the target within 5 seconds, or failed to hold the cursor at S and G, the trial was aborted and a 400~ms time window without reward was presented.

Note that this is a nontrivial reinforcement learning task, since the neurons did not ``know'' whether they belonged to the population C. Also, the network did not receive feedback about the cursor position, only binary information about the trial phase through the cue was provided. This is also true for the preferred directions assigned to the neurons in C, which could not be observed by the neurons. Furthermore, the neurons in C did not receive input from the cue directly, such that the routing of cue information to C had to be learned on top of the reaching task. All this information had to be discovered through random exploration from a global and sparse binary reward signal.

We used the synaptic sampling framework with and without the Hamiltonian momentum term to learn this task. Synaptic plasticity was here only active for excitatory synapses (both recurrent and feedforward), whereas inhibitory synapses were fixed. In order to guarantee that synapses didn't change their role, i.e., become inhibitory during learning, we used here a model for synaptic plasticity that does not allow synaptic weights to become negative. This was done by applying a mapping between synaptic parameters $\thi$ and the synaptic efficacies $w_i(t)$. We used here the exponential mapping
\begin{align}
 w_{i}(t) = \exp( \thi - \theta_0 )\;,\label{eq:thetamap}
\end{align}
with offset parameter $\theta_0=3$, such that $w_i(t)$ is positive for any value of $\thi$. We show in \nameref{sec:methods} that by inserting equation \eqref{eq:thetamap} into the general Hamiltonian learning framework \eqref{eq:sde}, we arrive at a slightly modified version of the eligibility trace $e_{i}(t)$, given by
\textbf{\begin{eqnarray}
  & \frac{de_{i}(t)}{dt} = -\frac{1}{\tau_e}\,~e_{i}(t) \;+\; w_{i}(t) \, \hz_{\prei}(t) \,  (\zkt -f_{\posti}(t)) \;. \label{eqn:eligibility-trace-mult-dyn} 
\end{eqnarray}}
This dynamics differs from equation \eqref{eqn:eligibility-trace} by the additional term $w_{i}(t)$, such that weight changes are scaled by the current value of the synaptic efficacy. This feature of our model mimics the multiplicative dynamics observed in cortical synaptic spines \cite{LoewensteinETAL:11}, see \cite{kappel2016reward} for a detailed analysis.

In Fig.~\ref{fig:Peter} we show that the Hamiltonian momentum term in the rule \eqref{eq:sde} significantly enhances learning this task. Network responses before and after learning with the Hamiltonian momentum term are shown in Fig.~\ref{fig:Peter}B. Initially the rewarded goal is only reached occasionally (around $20\%$ success rate, one example unsuccessful trial is shown). After learning for 12 hours the network is able to reach the target in most of the trials (success rate was $62\%$ on average, see Fig.~\ref{fig:Peter}C). In Fig.~\ref{fig:Peter}C we compare the learning progress with and without Hamiltonian sampling. We found that this task is hard to learn without the Hamiltonian momentum term (success rate typically below $40\%$ after 12 hours of learning).

\subsection{Hamiltonian dynamics improves network behavior at saddle points}

\begin{figure}
\begin{center}
\includegraphics[scale=0.9]{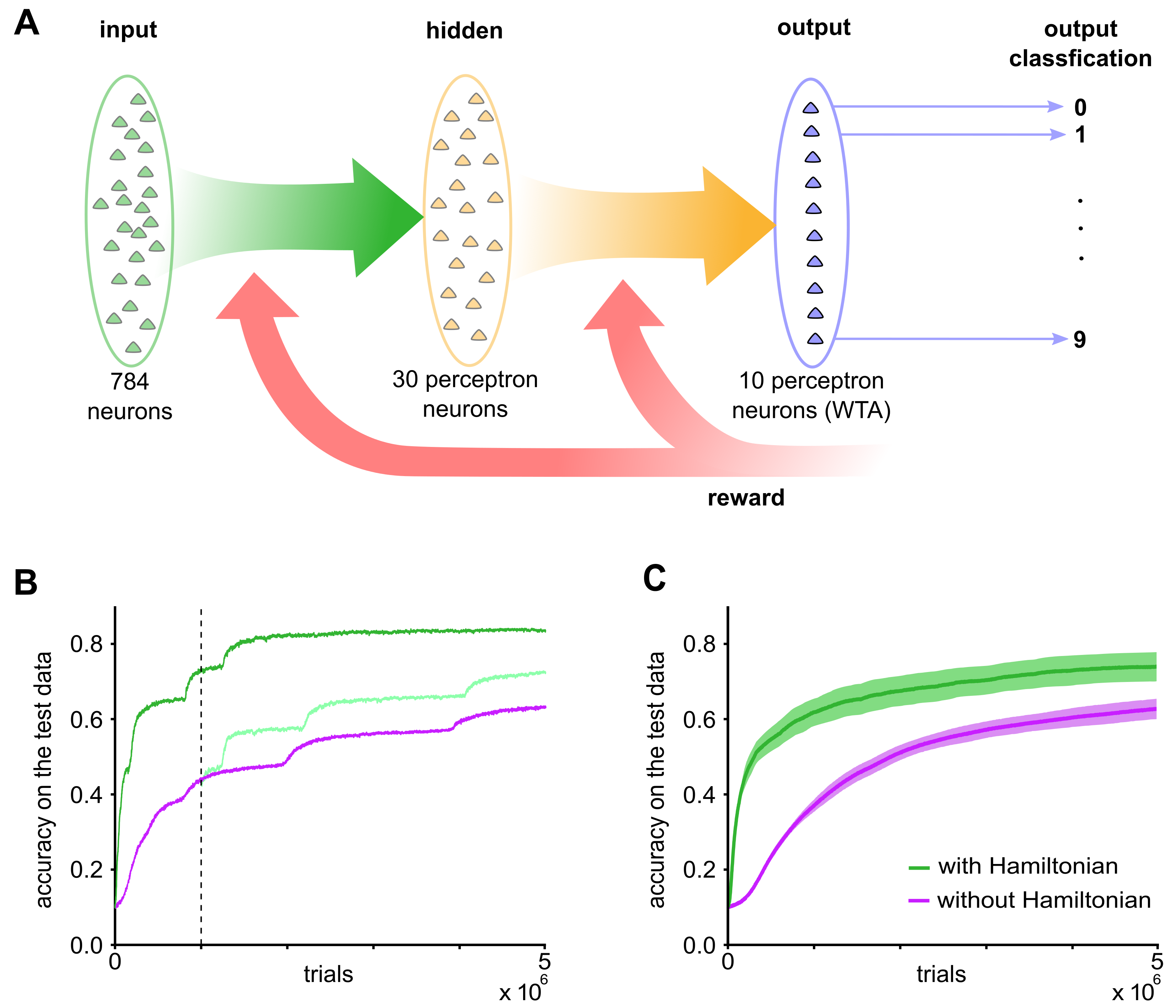}
\end{center}
\caption{{\bf Hamiltonian dynamics improves network behavior at saddle points.}
\pl{A} Network architecture.
\pl{B} Hamiltonian synaptic sampling can ease the saddle point problem. Network performance on the MNIST task during learning with Hamiltonian dynamics (dark green) and non-Hamiltonian synaptic sampling (magenta). With the same initial weights, Hamiltonian dynamics can escape saddle point more quickly. Dashed vertical line indicates time when non-Hamiltonian dynamics (magenta) was switched to Hamiltonian dynamics (light green) in the same network. 
\pl{C} Comparison of the average accuracy on the test data for Hamiltonian synaptic sampling (green) and non-Hamiltonian synaptic sampling (magenta; shading shows STDP over 30 independent learning trials).
} 
\label{fig:saddle}
\end{figure}

Traditionally, it was believed that gradient-based non-convex optimization in high-dimensional spaces is hampered by the presence of local optima in the fitness landscape. Recently, Dauphin et al. \cite{dauphin2014identifying,dauphin2015equilibrated} argued that in high-dimensional spaces there are typically only few local optima and that these local optima are nearly as good as the global optimum. Importantly, it was further noted by these authors that
saddle points are much more numerous in high-dimensional fitness landscapes.
Hence, stochastic procedures over high dimensional spaces, like synaptic sampling, tend to be inefficient and time consuming due to the presence of saddle points, but not so much due to local optima.
One generally accepted method to speed up convergence of learning or sampling in the presence of saddle points is to use Hamiltonian dynamics (or a momentum term). We therefore hypothesized that CaMKII-induced Hamiltonian parameter dynamics should provide a benefit in this respect. 
 
To test this hypothesis, we considered a three-layer neural network with 784 input neurons, 30 hidden neurons and 10 output neurons. The task was to learn to classify images of handwritten digits from the MNIST dataset (see Fig.~\ref{fig:saddle}A and \nameref{sec:methods}). Due to the large computational demands for this task, we did not consider a spiking network here but rather a network consisting of stochastic perceptrons (i.e., neurons with binary outputs which were set stochastically based on the weighted sum of inputs, similar to units in a Boltzmann machine). 
At each pattern presentation, one digit was chosen randomly from the MNIST dataset and presented as input. A binary reward was delivered depending on the activity of output neurons. If the output neuron corresponding to the target for the current example had larger firing probability than other output neurons, a reward of 1 was delivered, otherwise the reward was set to 0. Note that no eligibility trace was used as the network obtained feedback immediately (presentation of the pattern, computation of network output, and reward delivery were all performed in the same time step).

We first ran the network with non-Hamiltonian synaptic sampling (Fig.~\ref{fig:saddle}B, magenta curve).
The behavior of the network during learning showed typical signs of saddle points. In particular, the test accuracy tended to get stuck at some plateau value with only slight increases during longer periods. Then, at some point performance increased significantly (the network escaped from the saddle point) until another plateau was reached (see step-like behaivor of the magenta curve in Fig.~\ref{fig:saddle}B). Similar behavior was observed with Hamiltonian dynamics, however, in this case, the network tended to escape from saddle points much faster (Fig.~\ref{fig:saddle}B, green curve). To test whether Hamiltonian dynamics can escape saddle points faster than non-Hamiltonian synaptic sampling, we considered a parameter setting obtained by synaptic sampling close to a putative saddle point and continued the simulation with Hamiltonian dynamics (light green curve in Fig.~\ref{fig:saddle}B). We observed that the network escaped from the current saddle point much faster with the Hamiltonian dynamics. Considering the average performance for 30 independent learning trials, we found that Hamiltonian sampling accelerates learning significantly and obtains better result within reasonable learning times (Fig.~\ref{fig:saddle}C).

\subsection{From reward-based learning to global network optimization}

Virtually all previous approaches for reward-based learning in spiking neural networks are based on the policy gradient method, that is, the parameters of the network are gradually adjusted in the direction that increases the expected reward locally. Hence, for sufficiently long learning, the parameter setting of the network converges to a local optimum and stays at this local optimum thereafter. The proposed mathematical framework of Hamiltonian sampling allows us to create a link from reinforcement learning to nonlinear optimization theory and the simulated annealing algorithm. This link implies that (spiking or artificial) neural networks can in principle attain through learning not only functionally attractive locally optimal network configurations, but in principle even a global optimum. This theoretical result hence reveals a fundamental advantage of Hamiltonian synaptic dynamics over previous approaches for reward-based network optimization.

The link to nonlinear optimization becomes apparent when one takes a closer look at the temperature parameter $T$ in our plasticity dynamics \eqref{eq:sde} that scales the amount of noise in the parameter updates. Since for a given $T$, the network samples from $\pt = \frac{1}{\cal Z} p^*{\left( \bth  \right)^{\frac{1}{T}}}$, a decreased temperature $T<1$ concentrates parameter samples at values that lead to large rewards (for an uninformative prior) and therefore increases the expected reward of the network. In the limit $T\rightarrow 0$, the stationary distribution $\pt$ converges to the uniform distribution over optimal parameter settings with other parameter settings assuming zero probability 
\begin{equation}\label{eq:opt_distr}
    \lim_{T\rightarrow 0} \pt = \choosealt{ \frac{1}{|\mathcal{S}_{\text{opt}}|}}  {\bth \in \mathcal{S}_{\text{opt}}} 
    { 0 } { \bth \not \in \mathcal{S}_{\text{opt}}},
\end{equation}
where we have defined $\mathcal{S}_{\text{opt}}$ as the set of optimal network parameters and $|\mathcal{S}_{\text{opt}}| \equiv \int_{\bth \in \mathcal{S}_{\text{opt}}}  d\bth$ denotes the measure of this set, see \nameref{sec:methods}.
Further, the expected reward also assumes its global optimum in this limit. One attempt to attain such an optimum is to start with a large temperature and reduce it slowly towards 0. Such an annealing procedure is used in simulated annealing, a non-linear optimization technique \cite{dekkers1991global}. This cooling technique however needs convergence to the associated stationary distribution for each temperature $T$ within a reasonable time. 
While some data suggest that the genetic program for developmental learning has some features that are reminiscent of a cooling schedule \cite{GopnikETAL:15}, a Hamiltonian sampling dynamics is likely to improve the convergence speed for each temperature.

\begin{figure}
\begin{center}
\includegraphics[scale=1]{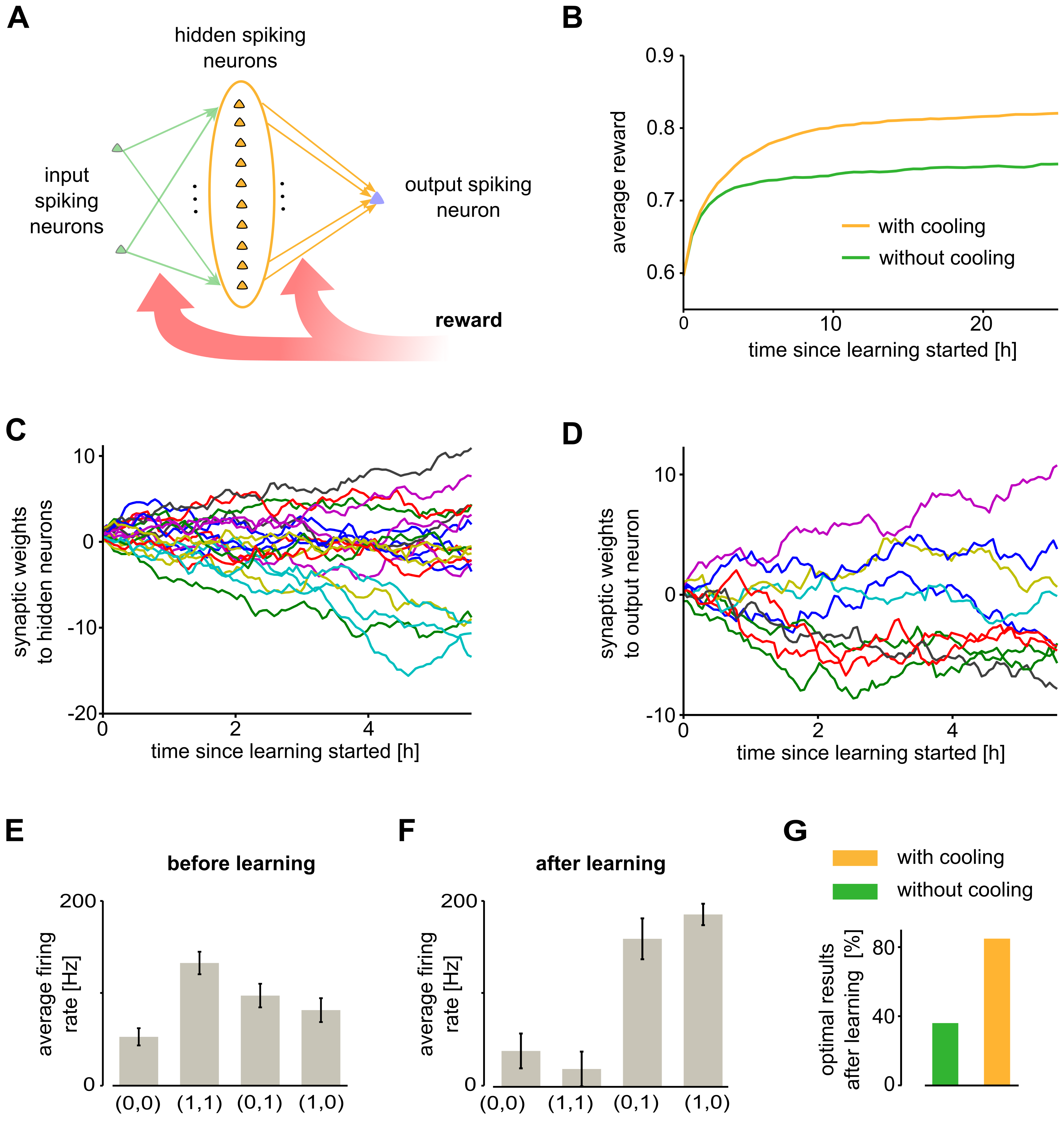}
\end{center}
\caption{{\bf Cooling improves reward-based learning in spiking neural networks.}
\pl{A} Illustration of the network architecture. The network consist of 2 input neurons, 10 hidden neurons and 1 output neuron. The task was to learn the XOR function.
\pl{B} Comparison of the average reward obtained during learning for Hamiltonian
dynamics with (orange) and without (green) cooling of the temperature $T$. 
\pl{C,D} Evolution of synaptic weights to neurons in the hidden layer (C) and to the output neurons (D) during the first 6 hours of learning.
\pl{E,F} Average firing rate of the output neuron for the four input patterns before (E) and after (F) learning.
\pl{G} Fraction of learning trials at which the network finds the optimal solution with (orange) and without (green) cooling.
} 
\label{fig:xor}
\end{figure}

We studied the benefit of a cooling schedule by considering a spiking neural network that learns the exclusive-or (XOR) function through reward-based learning (Fig.~\ref{fig:xor}A,B). The XOR function maps two binary variables to one binary output in the following manner: $\left(0,0\right) \to 0, \left(1,1\right) \to 0, \left(0,1\right) \to 1, \left(1,0\right) \to 1$. It is a classical task for artificial neural networks. The spiking neural network that we used for this task is shown in Fig.~\ref{fig:xor}A. It consisted of 2 input neurons, 10 hidden neurons and 1 output neuron. Each input neuron encoded one binary input variable. It produced a Poisson spike train at its output with a rate of $80$ Hz for the input $1$ and $3$ Hz  for the input $0$. The hidden neurons and output neuron were stochastic spiking neurons with a refractory time of $5$ ms. Each layer was fully connected to the next layer and initial synaptic weights were set randomly (see \nameref{sec:methods} for details).

During learning, a pattern was chosen randomly and presented to the network for $400$ ms. During this time, the output of the network was compared to the target output and a binary reward was delivered accordingly. More specifically, every $5$ ms the reward was recomputed and delivered to the network -- being $1$ if the output neuron spiked (was silent) in the past $5$ ms for a target of $1$ ($0$ respectively) and $0$ otherwise. The pattern presentation was followed by a delay period of $100$ ms where no input or reward was delivered to the network. Then, another randomly chosen pattern was presented and so on.

The evolution of the synaptic weights during learning is shown in Fig.~\ref{fig:xor}C,D. The weights of both layers change significantly throughout learning and contribute to learning the task. Synapses also remain plastic throughout the whole learning time and explore different solutions. Network responses before and after learning are shown in Fig.~\ref{fig:xor}E and Fig.~\ref{fig:xor}F. Before learning, the average firing rate of the output neuron for all the input patterns were 52.7, 132, 97.2, and 81.5 Hz respectively (see Fig.~\ref{fig:xor}E). After learning for 6 hours, the output neuron has maximized the firing rate for the input patterns $(0,1)$ and $(1,0$) and significantly reduced it for patterns $(0,0)$ and $(1,1)$ (see Fig.~\ref{fig:xor}F). 

This task was considered before by Seung and Xie \cite{seung2003learning,xie2004learning}. They also considered a stochastic spiking neuron model, however with zero refractory time. Further, in their model, positive or negative reward was delivered to the network every millisecond.
It was noted in \cite{xie2004learning} that learning does not work reliably if positive and negative rewards are not balanced. 
In fact using our highly unbalanced reward schedule with rewards being either $0$ or $1$ , the network often does not achieve optimal performance if a constant temperature is chosen for learning (Fig.~\ref{fig:xor}G). In this case, optimal results were obtained in only $40$ \% of the learning trials (where each trial was started with a different random initialization of the parameters). When we introduced a ``cooling'' schedule in which the temperature was decreased during learning, this ratio increased to $90$ \%.
The superiority of the annealed optimization is also visible in average reward attained during learning, see Fig.~\ref{fig:xor}B. This shows that parameter optimization with annealed noise can significantly improve performance of spiking neural networks. A similar observation was reported for deep artificial neural networks \cite{neelakantan2015adding}. Our theoretical framework of Hamiltonian sampling provides an explanation for this phenomenon as an optimization through annealed sampling similar to simulated annealing and thus opens the door to apply the toolkit of stochastic optimization to gradient-based neural network learning in a principled manner.

\section{Discussion}
We have presented a new theoretical framework for reward-based neural network optimization that integrates a hidden synaptic parameter in the plasticity process. We suggest that this synaptic parameter could be implemented in the synapse through \camk{}, that is abundantly present in the postsynaptic density and acts as low pass filter in the induction of synaptic plasticity. We have shown that the \camk{}-enriched dynamics supports a special type of ongoing stochastic policy search -- Hamiltonian sampling with friction -- and convergences to the stationary distribution much faster than Langevin sampling (synaptic sampling).

David Marr famously proposed to treat brain computation at three distinct, complementary levels of analysis  \cite{marr1976understanding}, which is today known  as Marr's Tri-Level Hypothesis. It is of interest to realize that biological data on the activation dynamics of the kinase \camk{}, corresponds to the implementational/physical level of Marr's Tri-Level Hypothesis. Our proposed model for network plasticity suggests that \camk{} enables the brain to perform Hamiltonian sampling on the algorithm level (algorithmic/representational level of Marr's Tri-Level Hypothesis). To be specific, biological networks of neurons are able to approximate Hamiltonian sampling of network configurations, rather than slower Langevin sampling or gradient descent.

We have demonstrated several advantages of Hamiltonian sampling over previously considered approaches to reward-based learning in spiking neural networks. We have shown in Fig.~\ref{fig:sigmoid} that this Hamiltonian synaptic sampling framework can be use to learn smooth responses of spiking neurons through reward-based learning, and that it further can scale up to learn recurrent networks of spiking neurons (shown in Fig.~\ref{fig:Peter}). In Fig.~\ref{fig:saddle} we have shown that synaptic sampling is prone to be slow near saddle points of the objective function, and that  Hamiltonian synaptic sampling can significantly speed up learning in these cases. Finally, we have demonstrated 
that reward-based network plasticity is in principle able to acquire through down-regulation of the stochastic component in parameter updates the full power of simulated annealing for optimizing the network. This allows neural networks to attain through learning not only locally optimal network configurations, but in principle even a global optimum. This theoretical result provides a new gold standard for reward-based network learning.

\camk{} dynamics has previously been studied in \cite{graupner_brunel_2007}. While this work focused on detailed molecular dynamics and its implications for STDP on the level of pairing protocols, we treated \camk{} dynamics in the current study more abstractly as a low-pass filtering process and studied the implications for system level reward-based learning. It is interesting to note that the low-pass filtering effect was also predicted in the model  of \cite{graupner_brunel_2007}. In addition they proposed a role of \camk{} for binary-state behavior of synapses in hippocampus. The underlying hypothesis that synaptic efficacies can attain only two possible states, a depressed state and a potentiated state, has been put into question by recent experimental data \cite{bartol2015nanoconnectomic}.

Our model makes a number of experimentally testable predictions. It was shown in previous work that synaptic spine dynamics can be modeled by a stochastic process (Ornstein-Uhlenbeck process) with two time-constants on the temporal scale of several days. Our model that includes the Hamiltonian momentum term suggests that also on short time scales (minutes to few hours), models of synaptic dynamics with two time constants should provide better fits. Moreover, the proposed role of \camk{} suggests that these time constant should correspond to rates of dephosphorylation.

Our result on network optimization in Fig.~\ref{fig:xor} suggests that biological networks are able to control the level of stochasticity, and that stochasticity decreases during long lasting learning processes (cooling).  Experimental results revealed that learning a new behavioral task is accompanied by increased synaptic spine numbers and spine dynamics \cite{PetersETAL:14,XuETAL:09}. In \cite{kappel2016reward} we analyzed a simple model for synaptic turnover and found that the statistics of spine regrowth during task acquisition can be explained by a brief increase of the learning temperature $T$. These findings suggest that the brain employs -- in addition to deterministic synaptic updates -- a mechanism to regulate the speed of random exploration in the high-dimensional space of synaptic parameters over several hours to days. This article has introduced a mathematical framework that provides a step towards understanding the complex interplay of deterministic and stochastic strategies employed by the brain, to solve complex learning problems.

\section{Methods}
\label{sec:methods}
\subsection{Details to learning the stationary distribution of network configurations through synaptic plasticity rules }
\label{sec:general}
Here we present the general mathematical framework of synaptic parameter dynamics and derive the emerging stationary distribution of network configurations that results from this dynamics. The generalized model, that includes both Hamiltonian synaptic sampling \eqref{eq:sde} and synaptic sampling without momentum \eqref{eq:sde_rmsynsam} as special cases, is given by the following set of SDEs:
\begin{equation}
\label{eqn:ham2}
\begin{array}{l}
d\theta _{i}(t) = \left( \left . - a \frac{\partial \log p^*( \bGamma )}{\partial \Gamma_{i}} \right|_{\bGamma(t)} + \left . c \frac{\partial \log p^*( \bth )}{\partial \theta _{i}} \right|_{\bth(t)} \right) dt + \sqrt {2 T c}~ d\mathcal{W}_{\theta_{i}}\\[2mm]
d\Gamma_{i}(t) = \left( \left . a \frac{\partial \log p^*(\bth)}{\partial \theta_{i}} \right|_{\bth(t)}  + \left . b \frac{\partial \log p^*( \bGamma )}{\partial \Gamma_{i}} \right|_{\bGamma(t)}  \right)dt + \sqrt {2T b}~ d\mathcal{W}_{\Gamma_{i}},
\end{array}
\end{equation}
where $p^*(\bth)$ is the posterior distribution over the network parameter given by equation \eqref{eq:posterior} and  $p^*(\bGamma)$ is the distribution over the \camk{}-related hidden synaptic parameter. The notation, $\left . \frac{\partial \log p^*( \bGamma )}{\partial \Gamma_{i}} \right|_{\bGamma(t)}$, denotes the derivative of $\log p^*( \bGamma )$ evaluated at the parameter vector $\bGamma(t)$. In \nameref{sec:results} we suppressed this time-dependences in order to simplify the notation.
$T>0$ is the temperature parameter, $\mathcal{W}_{\theta _{i}}, \mathcal{W}_{\Gamma _{i}}$ are independent one-dimensional Wiener processes, and $a$, $b$, $c$ are positive constants.

This dynamics describes a general noisy first-order interaction between visible synaptic parameters $\theta_{i}$, that determine the efficacy of the synapse, and hidden synaptic parameters $\Gamma_{i}$, the absolute value of which model the local concentration of \camk{} in its activated state. The dynamics can thus be seen as a generalization of standard gradient-based synaptic plasticity rules (e.g.~for maximum likelihood learning) that includes structural constraints, \camk{} activation and stochastic plasticity.
For the general dynamics, the joint distribution over the sets of all parameters $\bth$ and $\bGamma$ will converge after a while to $p^*_T(\bth, \bGamma) = \frac{1}{\cal Z} p^*( \bth )^{\frac{1}{T}} p^*( \bGamma )^{\frac{1}{T}}$ and produce samples from it. This result can be formalized in the following theorem:
\newtheorem{thm}{Theorem}[section]
\begin{thm}
	\label{lem:single_sup}
	Let $p^*({\bth })$,$p^*({\bGamma })$ be strictly positive, continuous probability distributions over parameters $\bth$ and $\bGamma$ respectively, twice continuously differentiable with respect to $\bth$ and $\bGamma$. Let $a,b,c$ be postitive constants. Then
	the set of stochastic differential equations (\ref{eqn:ham2}) leaves the distribution $p^*_T(\bth, \bGamma) =\frac{1}{{\cal Z}}p^*{\left( {\bth } \right)^{\frac{1}{T}}}p^*{\left( {\bGamma } \right)^{\frac{1}{T}}}$ invariant.
	Furthermore, this is the unique stationary distribution of the sampling dynamics.
\end{thm}

\emph{Proof}.
  Eq. (\ref{eqn:ham2}) has two drift terms ${A_{i}}\left( \bth  \right)$,   
 ${A_{i}}\left( \bGamma  \right)$ and two diffusion terms ${B_{{i},{s}}}\left( \bth \right)$, ${B_{{i},{s}}}\left( \bGamma  \right)$:
 
 \begin{equation}
 \label{eqn:drift1}
{A_{i}}\left( \bth  \right) = { - a\frac{{\partial \log p^*\left( {\bGamma  } \right)}}{{\partial {\Gamma _{i}}}} + c\frac{{\partial \log p^*\left( {\bth } \right)}}{{\partial {\theta _{i}}}} }
 \end{equation}
 
 \begin{equation}
\label{eqn:drift2}
 {A_{i}}\left( \bGamma  \right) = {a\frac{{\partial \log p^*\left( {\bth } \right)}}{{\partial {\theta _{i}}}} + b\frac{{\partial \log p^*\left( {\bGamma } \right)}}{{\partial {\Gamma _{i}}}} }
 \end{equation}

 \begin{equation}
 \label{eqn:diffusion1}
{B_{{i},{s}}}\left( \bth  \right) = \left\{ {\begin{array}{*{20}{c}}
\begin{array}{l}
2Tc ,i=s\\
\end{array}\\
{0,\;others}
\end{array}} \right.
 \end{equation}

 \begin{equation}
  \label{eqn:diffusion2}
{B_{{i},{s}}}\left( \bGamma  \right) = \left\{ {\begin{array}{*{20}{c}}
\begin{array}{l}
2Tb ,i=s\\
\end{array}\\
{0,\;others}
\end{array}} \right.
 \end{equation}
Hence the SDEs (\ref{eqn:ham2}) can be translate into the following Fokker-Planck equation:

\begin{equation}
\label{eqn:Fokker-Planck}
\begin{array}{l}
\frac{{d{p_{FP}}\left( {\bGamma ,\bth ,t} \right)}}{{dt}} = \sum\limits_{i} { - \frac{\partial }{{\partial {\theta _{i}}}}\left( {\left( { - a\frac{{\partial \log {p^*}\left( \bGamma  \right)}}{{\partial {\Gamma _{i}}}} + c\frac{{\partial \log {p^*}\left( \bth  \right)}}{{\partial {\theta _{i}}}}} \right){p_{FP}}\left( {\bGamma ,\bth ,t} \right)} \right)} \\
\;\;\;\;\;\;\;\;\;\;\;\;\;\;\;\;\;\;\;\;\;\;\;\;\; + \sum\limits_{i} { - \frac{\partial }{{\partial {\Gamma _{i}}}}} \left( {\left( {a\frac{{\partial \log {p^*}\left( \bth  \right)}}{{\partial {\theta _{i}}}} + b\frac{{\partial \log {p^*}\left( \bGamma  \right)}}{{\partial {\Gamma _{i}}}} } \right){p_{FP}}\left( {\bGamma ,\bth ,t} \right)} \right)\\
\;\;\;\;\;\;\;\;\;\;\;\;\;\;\;\;\;\;\;\;\;\;\;\;\; + \sum\limits_{i} {\frac{{{\partial ^2}}}{{\partial \theta _{i}^2}}} \left( {{T}c~{p_{FP}}\left( {\bGamma ,\bth ,t} \right)} \right) + \sum\limits_{i} {\frac{{{\partial ^2}}}{{\partial \Gamma _{i}^2}}} \left( {{T}b~{p_{FP}}\left( {\bGamma ,\bth ,t} \right)} \right),
\end{array}
\end{equation}
where $\frac{{d{p_{FP}}\left( {\bGamma ,\bth ,t} \right)}}{{dt}}$ denotes the distribution over network parameters at time $t$.
If we plug the stationary distribution ${p_T^ * }\left( {\bth ,\bGamma } \right) =\frac{1}{{\cal Z}} \left(p^*\left( {\bth } \right)p^*\left( {\bGamma } \right)\right)^{\frac{1}{T}}$ to the right side of eq.~\eqref{eqn:Fokker-Planck}, we have:
\begin{equation}
\begin{array}{l}
\frac{{d{p_{FP}}\left( {\bGamma ,\bth,t} \right)}}{{dt}}
= \sum\limits_{i} { - \frac{\partial }{{\partial {\theta _{i}}}}\left( {\left( { - a\frac{{\partial \log {p^*}\left( \bGamma  \right)}}{{\partial {\Gamma _{i}}}} + c\frac{{\partial \log {p^*}\left( \bth  \right)}}{{\partial {\theta _{i}}}} } \right) \frac{1}{{\cal Z}} \left(p^*\left( {\bth } \right)p^*\left( {\bGamma } \right)\right)^{\frac{1}{T}}   } \right)} \\
\;\;\;\;\;\;\;\;\;\;\;\;\;\;\;\;\;\;\;\;
+ \sum\limits_{i} { - \frac{\partial }{{\partial {\Gamma _{i}}}}} \left( {\left( {a\frac{{\partial \log {p^*}\left( \bth \right)}}{{\partial {\theta _{i}}}} + b\frac{{\partial \log {p^*}\left( \bGamma  \right)}}{{\partial {\Gamma _{i}}}}} \right) \frac{1}{{\cal Z}} \left(p^*\left( {\bth } \right)p^*\left( {\bGamma } \right)\right)^{\frac{1}{T}}  } \right)\\
\;\;\;\;\;\;\;\;\;\;\;\;\;\;\;\;\;\;\;\;
+ \sum\limits_{i} {\frac{{{\partial ^2}}}{{\partial \theta _{i}^2}}} \left( {Tc ~\frac{1}{{\cal Z}} \left(p^*\left( {\bth } \right)p^*\left( {\bGamma } \right)\right)^{\frac{1}{T}} } \right) + \sum\limits_{i} {\frac{{{\partial ^2}}}{{\partial \Gamma _{i}^2}}} \left( {Tb ~\frac{1}{{\cal Z}} \left(p^*\left( {\bth } \right)p^*\left( {\bGamma } \right)\right)^{\frac{1}{T}} } \right)\\
\;\;\;\;\;\;\;\;\;\;\;\;\;\;\;\;\;\;\;
= \sum\limits_{i} { - \frac{\partial }{{\partial {\theta _{i}}}}\left( {{c\frac{{\partial \log {p^*}\left( \bth  \right)}}{{\partial {\theta _{i}}}} } \frac{1}{{\cal Z}} \left(p^*\left( {\bth } \right)p^*\left( {\bGamma } \right)\right)^{\frac{1}{T}}  } \right)}  + \sum\limits_{i} {\frac{{{\partial ^2}}}{{\partial \theta _{i}^2}}} \left( {Tc ~\frac{1}{{\cal Z}} \left(p^*\left( {\bth } \right)p^*\left( {\bGamma } \right)\right)^{\frac{1}{T}} } \right) \\
\;\;\;\;\;\;\;\;\;\;\;\;\;\;\;\;\;\;\;\;
+ \sum\limits_{i} { - \frac{\partial }{{\partial {\Gamma _{i}}}}} \left( { {b\frac{{\partial \log {p^*}\left( \bGamma  \right)}}{{\partial {\Gamma _{i}}}}} \frac{1}{{\cal Z}} \left(p^*\left( {\bth } \right)p^*\left( {\bGamma } \right)\right)^{\frac{1}{T}} } \right)\; +\sum\limits_{i} {\frac{{{\partial ^2}}}{{\partial \Gamma _{i}^2}}} \left( {Tb ~\frac{1}{{\cal Z}} \left(p^*\left( {\bth } \right)p^*\left( {\bGamma } \right)\right)^{\frac{1}{T}} } \right)\\
\;\;\;\;\;\;\;\;\;\;\;\;\;\;\;\;\;\;\;
= \sum\limits_{i} {\frac{\partial }{{\partial {\theta _{i}}}}\left( { - c\frac{{\partial \log {p^*}\left( \bth  \right)}}{{\partial {\theta _{i}}}} \frac{1}{{\cal Z}} \left(p^*\left( {\bth } \right)p^*\left( {\bGamma } \right)\right)^{\frac{1}{T}}  + Tc \frac{1}{{\cal Z}} \left(p^*\left( {\bth } \right)p^*\left( {\bGamma } \right)\right)^{\frac{1}{T}}\frac{{\partial \log {p^*}{{\left( \bth  \right)}^{\frac{1}{T}}}}}{{\partial {\theta _{i}}}} } \right)  } \\
\;\;\;\;\;\;\;\;\;\;\;\;\;\;\;\;\;\;\;\;
+ \sum\limits_{i} {\frac{\partial }{{\partial {\Gamma _{i}}}}} \left( { - b\frac{{\partial \log {p^*}\left( \bGamma  \right)}}{{\partial {\Gamma _{i}}}} \frac{1}{{\cal Z}} \left(p^*\left( {\bth } \right)p^*\left( {\bGamma } \right)\right)^{\frac{1}{T}}  + Tb \frac{1}{{\cal Z}} \left(p^*\left( {\bth } \right)p^*\left( {\bGamma } \right)\right)^{\frac{1}{T}}  \frac{{\partial \log {p^*}{{\left( \bGamma  \right)}^{\frac{1}{T}}}}}{{\partial {\Gamma _{i}}}}  } \right)\\
\;\;\;\;\;\;\;\;\;\;\;\;\;\;\;\;\;\;\;
= 0
\end{array}
\end{equation}
This proves that $\frac{1}{{\cal Z}}p_T^*{\left( {\bth } \right)^{\frac{1}{T}}}p^*{\left( {\bGamma } \right)^{\frac{1}{T}}}$ is the stationary distribution of the parameters dynamic (\ref{eqn:ham2}). Under the assumption that $b$ and $c$ are strictly positive, this stationary distribution is also unique.
 If the matrix of diffusion coefficients is invertible, and the potential conditions are satisfied, the stationary distribution can be obtained (uniquely) by simple integration.  Since the matrix of diffusion coefficients is diagonal in our model, the diffusion
coefficient matrix is trivially invertible if all diagonal elements, i.e. all $b$ and $c$ are strictly positive. Also the potential conditions are fulfilled (by design), as can be verified by substituting eqs.~(\ref{eqn:drift1} -- \ref{eqn:diffusion2}) into Equation (5.3.22) in \cite{gardiner2004handbook},

\begin{align}
Z_{\theta_{i}}\left( \bth,\bGamma \right)~&=~ {B_{{i},{i}}^{-1}}\left( \bth  \right)  \left( 2A_{i}(\bth)-\frac{\partial}{\partial \theta_{i}}B_{{i},{i}}(\bth) \right) \nonumber\\
~&=~\frac{1}{2Tc}\left(  { - 2a\frac{{\partial \log p^*\left( {\bGamma  } \right)}}{{\partial {\Gamma _{i}}}} + 2c\frac{{\partial \log p^*\left( {\bth } \right)}}{{\partial {\theta _{i}}}} }   \right)
\end{align}

\begin{align}
Z_{\Gamma_{i}}\left( \bth,\bGamma \right)~&=~ {B_{{i},{i}}^{-1}}\left( \bGamma \right)  \left( 2A_{i}(\bGamma)-\frac{\partial}{\partial \Gamma_{i}}B_{{i},{i}}(\bGamma) \right) \nonumber\\
~&=~\frac{1}{2Tb}\left( {a\frac{{\partial \log p^*\left( {\bth } \right)}}{{\partial {\theta _{i}}}} + b\frac{{\partial \log p^*\left( {\bGamma } \right)}}{{\partial {\Gamma _{i}}}} } \right)
\end{align}
This shows that $ Z(\bth,\bGamma)=\left(Z_{\theta_{i}}(\bth,\bGamma), Z_{\Gamma_{i}}(\bth,\bGamma) \right)$ is a gradient. Thus, the potential conditions are met and the stationary distribution is unique.

In Theorem \ref{lem:single_sup}, $b$,$c$ need to be strictly positive. Note that we can relax it to $b$ or $c$ is strictly positive (or both) -- which means there exists diffusion noise -- and can prove $\frac{1}{{\cal Z}}p_T^*{\left( {\bth } \right)^{\frac{1}{T}}}p^*{\left( {\bGamma } \right)^{\frac{1}{T}}}$ is a unique stationary distribution of stochastic differential equations (\ref{eqn:ham2}) in the same way.  

Hamiltonian synaptic sampling \eqref{eq:sde}
and synaptic sampling \eqref{eq:sde_rmsynsam} are special cases of the more general parameter dynamics \eqref{eqn:ham2}.
Hamiltonian synaptic sampling as defined in (\ref{eq:sde}) is obtained by choosing  $c = 0$ and a Gaussian distribution for the hidden parameters $p^*({\bGamma })\sim NORMAL(0,1)$. Synaptic sampling as defined in (\ref{eq:sde_rmsynsam}) is obtained by choosing $a=b=0$. 
We remark that various types of gradient descent can also be recovered from the generalized dynamics for $T = 0$, e.g. gradient descent with momentum for the noiseless Hamiltonian dynamics.
 Equation (\ref{eqn:ham2}) can be seen as the continuous version of Hamiltonian sampling \cite{neal2011mcmc}, where a Metropolis update is performed after simulating Hamiltonian dynamic.  
Equation (\ref{eqn:ham2}) can also be seen as an extension of stochastic gradient Hamiltonian  Monte Carlo with friction \cite{chen2014stochastic, ma2015complete} to the case where the temperature $T$ is used to shape the static distribution $\pt$.

\subsection{Spiking neuron model}
\label{sec:spiking-neuron-model}
Spiking neurons were modeled by a stochastic variant of the spike response model \cite{gerstner2014neuronal}.
We use $w_{i}(t)$ to donate the synaptic efficacy of the $i$-th synapse in the network at time $t$. 
Each neuron $z_k$ of network $\mathcal{N}$ is then modeled as a point neuron with membrane potential $u_k(t)$ at time $t$
\begin{align}
u_k(t) &\;=\; \sum_{i \in  \syn{k}} \hz_{\preidef}(t)\, w_{i}(t) \;+\; \varphi_k(t) \; , \label{eq:membrane-potential}
\end{align}
where $\syn{k}$ is the index set of synapses that project to neuron $z_k$,  $\preidef$ denote the index of the presynaptic neuron of synapse $i$, $\varphi_k(t)$ denotes the bias potential of neuron $z_k$. In the recurrent network in Fig.~\ref{fig:Peter} we used a slowly changing bias potential to ensures that the output rate of each neuron stays within finite bounds (described in detail below). In all other experiments we used a constant bias potential. $\hz_{\preidef}(t)$ denote the trace of the (unweighted) postsynaptic potentials (PSPs) from  presynaptic neuron of synapse $i$ at time $t$. Throughout this paper, we used standard double-exponential PSP kernels with a brief finite rise and exponential decay, of the form $\epsilon(t) = \frac{\tau_r}{\tau_m - \tau_r} \left( e^{-\frac{t}{\tau_m}} - e^{-\frac{t}{\tau_r}}  \right)$, with time constants $\tau_m$ and $\tau_r$ (any other PSP shape may be used in principle without further adaptations of the theoretical model).

We denote the output spike train of neuron $z_k$ by $z_k(t)$, which is defined as a sum of Dirac delta pulses positioned at the spike times $t_k^{(1)}, t_k^{(2)}, \dots$, i.e., $z_k(t) = \sum_l \delta(t-t_k^{(l)})$. Neuron fires according to the link function $f_k(t)$ which denotes the firing probability of neuron $k$ at time $t$. Due to the lasting effects of PSPs, the firing probability may depend on the history of past spiking activities of all $K$ input neurons up to time $t$ which we denote by $\bbxt = \{x_i(\tau) \;\vert\; 1 \leq i \leq K, 0 \leq \tau < t \}$, which is defined as:
\begin{equation}\label{eqn:fire-prob}
\pn{z_k(t)=1}{\bbxt,\bth} =\; f_k(t) \;=\; f(u_k(t), \rho_k(t))\; ,
\end{equation}
where $\rho_k(t)$ denotes a refractory variable that is given by the time elapsed since the last spike of neuron $z_k$. In this article, we set
$f(u_k, \rho_k) = \sigma(u_k) \Theta(\rho_k-t_{ref})$, where $\sigma(u_k)$ is a sigmoid activation function  $\sigma(u_k) = \frac{1}{1+ \,e^{-u_k}}$ and $\Theta(\cdot)$ denotes the Heaviside step function, i.e. $\Theta(x)=1$ for $x\ge 0$ and $0$ otherwise. In our simulation, we set refectory time $t_{ref}$ to 5 ms.

\subsection{Reward-modulated synaptic plasticity rule}
Here we derive the reward-based learning rules for the spiking neural network model outline above.
In particular, we compute here the gradient of the expected reward:
\begin{equation}
\label{eq:gradexp}
\ddthetai\log\mathcal{V}(\bth) \;=\; \ddthetai\log   \expect[p(\ve r | \bth)]{ \int_{0}^\infty e^{-\frac{\tau}{\tau_e}} \,r(\tau) \; \d \tau } \;.
\end{equation}
We only consider the recurrent network in section \nameref{sec:peters} as an example and show that the parameter dynamic (\ref{eqn:eligibility-trace}), (\ref{eqn:gradient-est}) approximate this gradient. Actually one can compute the gradient of the expected reward for the feed-forward neural network in other simulations and get similar learning rule.
In order to simplify notation, we use $\bbzt$ to represent the history of past spiking activity of all neurons $z_k(1 \leq k \leq K)$ up to time t. Supposing that the reward signal $r(\tau)$ is only decided by $\bbzt$, we can rewrite $\ddthetai\log\mathcal{V}(\bth)$ as the expectation over all possible spike trains $\bbzt$ up to time t:
\begin{eqnarray}
\ddthetai \log \mathcal{V}(\bth)  &\;=\;& \frac{1}{\mathcal{V}(\bth)}  \expect[p(\ve r, \bbz | \bth)]{ \int_{0}^\infty  e^{-\frac{\tau}{\tau_e}} \, r(\tau) \,
                                                         \ddthetai \log \cprob{r(\tau), \bbztau}{\bth}  \; \d \tau } \;
 \nonumber \\[3mm]
  &\;=\;&  \expect[p(\ve r, \bbz | \bth)]{ \int_{0}^\infty  e^{-\frac{\tau}{\tau_e}} \,  \frac{r(\tau)}{\mathcal{V}(\bth)} \, \ddthetai \log \pn{ \bbztau}{\bth} \; \d \tau } \;.
\label{eqn:derivation-dlogpndtheta-1}
\end{eqnarray}
Note that we use the fact $ \log \cprob{r(\tau), \bbztau}{\bth}=\log \cprob{r(\tau)}{\bbztau} +\log \cprob{\bbztau}{\bth}$. The problem now is to estimate the gradient of the probability of observing the spike train $\bbztau$ in the time interval $0$ to $\tau$. According to Eqs. \eqref{eq:membrane-potential} and \eqref{eqn:fire-prob}, the logarithm of the probability distribution $\pn{\bbztau}{\bth}$ can be rewritten as:

\begin{equation}
 \log \pn{\bbztau}{\bth} \;=\; \int_0^t  \big( \zks \, \log \fks \,-\, (1-\zks) \,\log ( 1 - \fks) \big) \d s
  \label{eqn:network-spike-prob}
\end{equation}
where the integration runs from time $0$ to $t$. Using this, the gradient $\ddthetai \log \pn{\bbztau}{\bth}$ can be estimated
\begin{align}
   \ddthetai \log \pn{\bbztau}{\bth} \;&=\; \int_0^\tau \frac{\partial w_{i}}{\partial \theta_i} \dd{w_{i}} 
    \big( \zks \, \log \fks \,-\, (1-\zks) \,\log ( 1 - \fks) \big) \, \d s \nonumber \\[3mm]
     \;&=\;  
     \int_0^\tau w_{i} \, \hz_{\prei}(s) \, (\zks-\fks) \, \d s \;. \label{eq:ddw_sigm}
\end{align}
The dependence on $w_i$ (the current value of the synaptic weight), is a result of applying the chain rule and using the exponential mapping function \eqref{eq:thetamap}. If a linear mapping between $\theta_i$ and $w_i$ is used this term vanishes as in Eq.~\eqref{eqn:eligibility-trace}. The learning rules are similar to previous ones which were found in the context of maximum likelihood and reinforcement learning in neural networks \cite{pfister2006optimal, brea2013matching}.

Eq. ~\eqref{eqn:derivation-dlogpndtheta-1} defines a batch learning rule with an average taken over learning episodes where in each episode network responses and rewards are drawn according to the distribution $p(\ve r, \bbz | \bth)$.
In order to arrive at an online learning rule for this scenario, we consider an estimator of Eq.~\eqref{eqn:derivation-dlogpndtheta-1} that approximates its value at each time $t>\tau_g$ based on the recent network activity and rewards during time $[t-\tau_g, t]$ for some suitable $\tau_g>0$. We denote the estimator at time $t$ by $G_i(t)$ where we want $G_i(t) \approx \ddthetai \log \mathcal{V}(\bth)$ for all $t>\tau_g$. To arrive at such an estimator,
we approximate the average over episodes in Eq.~\eqref{eqn:derivation-dlogpndtheta-1} by an average over time where each time point is treated as the start of an episode. The average is taken over a long sequence of network activity that starts at time $t$ and ends at time $t+\tau_g$. Here, one systematic difference to the batch setup is that one cannot guarantee a time-invariant distribution over initial network conditions as we did there since those will depend on the current network parameter setting. However, under the assumption that the influence of initial conditions (such as initial membrane potentials and refractory states) decays quickly compared to the time scale of the environmental dynamics, it is reasonable to assume that the induced error is negligible.
We thus rewrite Eq.~\eqref{eqn:derivation-dlogpndtheta-1} in the form
\begin{equation*}
 \ddthetai \log \mathcal{V}(\bth) \;\approx\; G_i(t) =  \frac{1}{\tau_g} \int_{t}^{t+\tau_g} \int_{\zeta}^{t+\tau_g} \; e^{-\frac{\tau-\zeta}{\tau_e}} \;  \frac{r(\tau)}{\mathcal{V}(\bth)} \; \int_{\zeta}^{\tau} \, w_{i}(s) \, \hz_{\prei}(s) \, (\zks\,-\,\fks))  \; \d s \; \d \tau  \; \d \zeta  \;, 
\end{equation*}
where  $\tau_g$ is the length of the sequence of network activity over which the empirical expectation is taken. 
Finally, we can combine the second and third integral into a single one, rearrange terms and substitute $s$ and $\tau$ so that integrals run into the past rather than the future, to obtain
\begin{equation}
  G_i(t) \;\approx\; \frac{1}{\tau_g} \int_{t-\tau_g}^{t} \; \frac{r(\tau)}{\mathcal{V}(\bth)} \; \int_{0}^{\tau} e^{-\frac{s}{\tau_e}} \, w_{i}(\tau-s) \, \hz_{\prei}(\tau-s) \, (\zk(\tau-s)\,-\,\fk(\tau-s))  \; \d s \; \d \tau \;.
 \label{eqn:prthetadtheta}
\end{equation}
Supposing that $\tau_g$ tends to $0$, we get a simple on-line learning rule to approximate $G_i(t)$:
\begin{eqnarray}
  &  G_i(t) \approx r(t) \, e_{i}(t) \;. \label{eqn:gradient-est1}\\
  & \frac{de_{i}(t)}{dt} = \left(-\frac{1}{\tau_e}\,~e_{i}(t) \;+\; w_{i}(t) \, \hz_{\prei}(t) \,  (\zkt -f_{\posti}(t))\right) \label{eqn:eligibility-trace1} 
\end{eqnarray}
A similar learning rule has already been proposed by Seung and Xie \cite{xie2004learning}. In fact, as the learning rule only estimate Eq. \eqref{eqn:prthetadtheta} based on the reward at time $t$, it ignores outer integral in Eq. \eqref{eqn:prthetadtheta} and thus can't approximate $G_i(t)$ accurately. A better estimation has been given by Kappel et al. \cite{kappel2016reward} to improve learning performance, but without biological plausible motivation. Actually in our Hamiltonian synaptic sampling framework, \camk{} works as a momentum term that computes the average of the gradient  $\ddthetai \log \mathcal{V}(\bth)$ instead of current gradient during on-line learning, which corresponds to the outer integral and thereby supporting better estimate of  the gradient of expected reward.

\subsection{Relating Hamiltonian synaptic sampling to synaptic sampling}
Here we build the relationship between Hamiltonian synaptic sampling and synaptic sampling and show that synaptic sampling is included in Hamiltonian synaptic sampling. 
For simplicity and brevity, here we consider a version of the parameters dynamics for discrete time. According to Eq. \eqref{eq:sde_rmsynsam}, the parameter change $\Delta \theta_{i}^{syn}$ of synaptic sampling during a small discrete time step $\Delta t$ can be written as:
\begin{equation}
\label{eq:dis_syn_sampling}
  \Delta \theta_{i}^{syn} \;=\; \beta \, \Delta t \, \ddthetai \, \log p^*( \bth)  \;+\; \sqrt {2T\beta \Delta t}~ v_{i}^t~,
\end{equation}
where $\beta>0$ denotes a learning rate that controls the speed of the parameter dynamics.$v_{i}^t$ represents Gaussian noise with zero mean and variance 1. These noises are independent for each parameter $\theta_{i}$ and each update time $t$.

To compare synaptic sampling with Hamiltonian synaptic sampling, we rewrite Eq. (\ref{eq:sde}) into the discrete version with the same time step $\Delta t$:
\begin{equation}
\label{eq:dis_SDE}
\begin{array}{l}
 \Delta \theta_{i}^{ham} \;=\; a~\Delta t~\Gamma_{i}(t+\Delta t ),\\
  \Gamma_{i}(t+\Delta t )\; =\; (1-b~\Delta t)~ \Gamma_{i}(t)  \;+\; b~\Delta t \left( \frac{a}{b} \, \ddthetai \log p^*( \bth) \;+\;     \sqrt { \frac{2T}{b\Delta t} }~ v_{i}^t~ \right).\\
\end{array}
\end{equation}
Eq. (\ref{eq:dis_SDE}) seems different from Eq. (\ref{eq:dis_syn_sampling}). Actually, we can build the relationship between Eq. (\ref{eq:dis_SDE}) and (\ref{eq:dis_syn_sampling}) with the assumption that the momentum term $\Gamma_{i}$ has transient time constant (tends to zeros). To be specific, the parameter $b$ is very large and $b\Delta t$ tends to be 1. We thus rewrite the discrete version of Hamiltonian synaptic sampling (\ref{eq:dis_SDE}) as:
\begin{equation}
\label{eq:re_dis_SDE}
  \Delta \theta_{i}^{ham}  \;=\; \frac{a^2}{b} ~\Delta t \, \ddthetai~ \log p^*( \bth)  \;+\; \sqrt {\frac{2T a^2\Delta t}{b}}~ v_{i}^t~.
\end{equation}
Note that Eq. (\ref{eq:re_dis_SDE}) and (\ref{eq:dis_syn_sampling}) are the same if $\frac{a^2}{b}=\beta$ holds. Hence we conclude that synaptic sampling is a special case of Hamiltonian synaptic sampling that the momentum term changes on transient time constant. 

\subsection{Global network optimization through stochastic synaptic plasticity} \label{sec:expected_reward}

Here, we show that in principle, stochastic plasticity with a cooling schedule (i.e., with a slow decrease of the noise amplitude) can produce globally optimal network configurations.

{\bf Temperature-dependent expected reward:} We first calculate the expected reward that is attained by a network with parameters that have converged to the stationary distribution $\pt = \frac{1}{\mathcal{Z}}p^*(\bth|R=1)^\frac{1}{T}$ at temperature $T$. We denote by $R_T$ the Bernoulli random variable that indicates reward at temperature $T$.
When the network has reached the stationary distribution $\pt$, the expected reward $E[R_T]$ is given by
\begin{equation}
\begin{split}
  E[R_T] &= \sum_{r\in\{0,1\}} r~ p_\mathcal{N}(R_T=r) = p_\mathcal{N}(R_T=1) = \nonumber \\
       &= \int \pn{R=1}{\bth} \pt d\bth = \frac{1}{\mathcal{Z}} \int \pn{R=1}{\bth} p^*(\bth|R=1)^{\frac{1}{T}} d\bth \nonumber \\
       &= \frac{1}{\mathcal{Z}} \int \pn{R=1}{\bth} \frac{\pn{R=1}{\bth}^{\frac{1}{T}} \ps{\bth}^{\frac{1}{T}}}{p_\mathcal{N}(R=1)^{\frac{1}{T}}} d\bth.
\end{split}
\end{equation}
Assuming an uninformative prior $\ps{\bth}$, we obtain
\begin{equation}
\begin{split}
  E[R_T] &= \frac{1}{\mathcal{Z}_T} \int \pn{R=1}{\bth} \pn{R=1}{\bth}^{\frac{1}{T}} d\bth \nonumber = \frac{1}{\mathcal{Z}_T} \int \pn{R=1}{\bth}^{1+\frac{1}{T}} d\bth,
\end{split}
\end{equation}
where $\mathcal{Z}_T=\int \pn{R=1}{\bth}^{\frac{1}{T}} d\bth$ normalizes $\pn{R=1}{\bth}^{\frac{1}{T}}$ with respect to $\bth$. In other words, sampling from the posterior $p^*(\bth)$ with decreased temperature $T<1$ concentrates parameter samples at values that lead to large rewards and therefore increases the expected reward of the network.

{\bf Temperature annealing for global optimization:}
For small temperatures, the posterior is concentrated at the global optimum of the reward landscape.
In practice, the sampling process mixes extremely slowly at low temperatures due to low probabilities for non-optimal states. Hence, an annealing schedule that decreases the temperature slowly over time has to be employed in order to give synaptic sampling enough time to settle at the global optimum, similar to the simulated annealing optimization algorithm.

Here we show that in the limit $T\rightarrow 0$, the network achieves
maximal possible performance (the derivation is similar to the one in \cite{aarts1988simulated} for simulated annealing). Let $\mathcal{S}_{\text{opt}}$ denote the
set of optimal circuit parameters, i.e., $\mathcal{S}_{\text{opt}} = \{ \bth \, | \, \pn{R=1}{\bth} = R_{\text{max}}\}$ for $R_{\text{max}} \equiv \max_{\bth} \{\pn{R=1}{\bth}\}$.
For an uninformative prior we have
\begin{eqnarray}
  \lim_{T\rightarrow 0} \pt = \lim_{T\rightarrow 0} \frac{\pn{R=1}{\bth}^{\frac{1}{T}}}{\mathcal{Z}_T} &=& \lim_{T\rightarrow 0} \frac{\pn{R=1}{\bth}^{\frac{1}{T}}}{\int \pn{R=1}{\bth}^{\frac{1}{T}} d\bth} \\
  &=& \lim_{T\rightarrow 0} \frac{R_{\text{max}}^{-\frac{1}{T}} \pn{R=1}{\bth}^{\frac{1}{T}}}{\int R_{\text{max}}^{-\frac{1}{T}} \pn{R=1}{\bth}^{\frac{1}{T}} d\bth} \\
  &=& \lim_{T\rightarrow 0} \frac{\left(\frac{\pn{R=1}{\bth}}{R_{\text{max}}}\right)^{\frac{1}{T}}}{\int \left(\frac{\pn{R=1}{\bth} }{R_{\text{max}}}\right)^{\frac{1}{T}} d\bth}.
\end{eqnarray}
This evaluates to
\begin{equation}
    \lim_{T\rightarrow 0} \pt = \lim_{T\rightarrow 0} \frac{\pn{R=1}{\bth}^{\frac{1}{T}}}{\mathcal{Z}_T} = \choosealt{\frac{1}{\int_{\bth \in \mathcal{S}_{\text{opt}}}  d\bth} = \frac{1}{|\mathcal{S}_{\text{opt}}|}}  {\bth \in \mathcal{S}_{\text{opt}}} {
   0 } { \bth \not \in \mathcal{S}_{\text{opt}}},
\end{equation}
where we have defined $|\mathcal{S}_{\text{opt}}| \equiv \int_{\bth \in \mathcal{S}_{\text{opt}}}  d\bth$.
Hence, in the limit $T\rightarrow 0$, the distribution is a uniform distribution over optimal parameter values. For the expected reward, we thus have
\begin{eqnarray}
  \lim_{T\rightarrow 0} E[R_T] = \int_{\bth \in \mathcal{S}_{\text{opt}}} R_{\text{max}} \frac{1}{|\mathcal{S}_{\text{opt}}|} d\bth = R_{\text{max}}.
\end{eqnarray}

\subsection{Simulation details and parameters }
\vspace{-5mm}
\subheading{Details to: A spiking neuron learns to emulate a sigmoidal neuron (Fig.~\ref{fig:sigmoid})}
The network architecture is shown in Fig.~\ref{fig:sigmoid}A. The sigmoidal neuron receives inputs from 4 input numbers with the pre-defined weights $4,3,-3,-6$ and basis $1$. The spiking neuron receives inputs from 80 Poisson spiking neurons, which are divided into 4 pools. Each pool of 20 neurons encodes the same input number.
In order to generate the input patters, we first generate 2000 random vectors with dimension 4 by sampling from a uniform distribution on $[0,1]$, and then choose 20 vectors as input. Fig.~\ref{fig:sigmoid}B shows how the 20 examples are distributed along the input (i.e., the weighted sum of the four inputs) - output plane of a sigmoidal neuron. For each node, the X-coordinate represents the weighted sum of the four input numbers and the Y-coordinate represents the output of the sigmoidal neuron. The mapping between $x$ and $y$ is defined as $y=\frac{1}{1+e^{-x}}$ for this sigmoidal neuron. 

At each presentation, one out of the 20 input patterns is chosen randomly as input, which is converted to the Poisson spike trains with space-rate coding. The maximum firing rate of each neuron is set to 60 HZ.  After presenting the example for 300~ms, a reward is delivered to the network for 10 ms. 
The reward amplitude is given by computing the difference between the output of the sigmoidal neuron and scaled firing rate of spiking neuron, that is, $r=1-\left| f_1-f_2\right|$, where $f_1$ denotes the firing probability of the sigmoidal neuron , $f_2$ denotes the scaled firing rate (average firing probability) of the spiking neuron during the 300 ms time window of pattern presentation. Note that $f_2$ equals to the rate of the total firing times of the spiking neuron during the 300 time window to the maximum firing times 60 (as the refectory time is 5 ms). The time constants for the eligibility trace and the momentum are set to 0.2 s and 10 s respectively.

\subheading{Details to: Reward-guided network plasticity (Fig.~\ref{fig:Peter})}
The network connectivity between excitatory and inhibitory neurons was as suggested in \cite{AvermannETAL:12}. Excitatory and Inhibitory neurons were randomly connected with connection probabilities  given as in Table~2 of \cite{AvermannETAL:12}. Connections include lateral inhibition between excitatory and inhibitory neurons. The connectivity to and from inhibitory neurons was kept fixed (not subject to learning). The connection probability from excitatory to inhibitory neurons was given by 0.575. The synaptic weights were drawn from a Gaussian distribution (truncated at zero) with $\mu = 1$ and $\sigma = 0.1$. Inhibitory neurons were connected to their targets with probability $0.6$ (to excitatory neurons) and 0.55 (to inhibitory neurons) and the synaptic weights were drawn from a truncated Gaussian distribution with $\mu = -2$ and $\sigma = 0.2$. The connectivity from and to inhibitory neurons is kept fixed throughout the simulation. Plastic synaptic connections were allowed between all pairs of input and excitatory hidden neurons and among excitatory hidden neurons. The number of potential excitatory synaptic connections between each pair of neurons was drawn from a Binomial distribution ($p=0.5$, $n=10$).

The refractory period was 5~ms for excitatory and 2~ms for inhibitory neurons. For the postsynaptic potentials $y_i(t)$ of excitatory neurons, we used time constants $\tau_m = 20~\text{ms}$ and $\tau_r = 2~\text{ms}$. For inhibitory synapses we used a faster kernel of the same form with $\tau_m = 10~\text{ms}$ and $\tau_r = 1~\text{ms}$. The bias potential $\vartheta_k(t)$ in Eq.~\eqref{eq:membrane-potential} was initialized at -3 and then followed the dynamics
\begin{equation}
  \tau_{\vartheta}\, \frac{\d \vartheta_k(t)}{\d t} \;=\; \nu_0 - z_k(t)  \;,
  \label{eqn:adaptation-current}
\end{equation}
where $\tau_{\vartheta} = 50~\text{s}$ is the time constant of the adaptation mechanism and $\nu_0 = 5~\text{Hz}$ is the desired output rate of the neuron. Eq.~\eqref{eqn:adaptation-current} is a simplified version of the mechanism proposed in \cite{RemmeWadman:12} to balance activity in networks of excitatory and inhibitory neurons. We found that this regularization significantly increased the performance and learning speed of our network model.

For the synaptic dynamics we used a Gaussian prior $\ps{\bth}$ with $\mu=0$ and $\sigma=2$ and synaptic parameters were initially drawn form a Gaussian distribution with $\mu=-0.5$ and $\sigma=0.5$. Synaptic parameter changes were clipped at $\pm 40$ and synaptic parameters were not allowed to exceed $[-2,5]$ for the sake of numerical stability. The weights of synapses for which the synaptic parameters $\theta_i(t)$ became smaller than zero, where clamped to $w_i(t)=0$ as in our previous model \cite{kappel2016reward}. The temperature parameter $T$ was kept here constant at $T=0.1$. The time constants for the eligibility trace and the momentum were set here to 1~s and 50~s, respectively.

To generate the cue pattern we adapted the method from \cite{kappel2015network}. The afferent inputs were given here by representations of a simple symbolic sensory environment. Inputs were randomly generated realizations of inhomogeneous Poisson spike trains. To generate these spike patterns,  each of the 200 input neurons  was assigned to a Gaussian tuning curve with $\sigma=0.2$. Tuning curve centers were independently and equally scattered over the unit cube. The cue was represented by a randomly selected point in this 3-dimensional space which is covered by the tuning curves of the input neurons. The stimulus positions was overlaid with small-amplitude jitter ($\sigma=0.05$). For each presentation of the cue the firing rate of an individual input neuron was given by the support of sensory experience under the input neuron's tuning curve. The maximum firing rate of each input neuron was 60~Hz.  In addition an offset of 2~Hz background noise was added. If no cue was present, all input neurons were set to homogeneous 2~Hz Poisson firing.

The preferred direction for the population vector decoding of each neuron was drawn independently for the X- and Y-axis in cursor space from a uniform distribution in the interval $\pm 0.025$. Cursor movement was implemented using a simple version of the population vector method \cite{georgopoulos1986neuronal}. Each spike of a neuron in C caused the cursor to jump into the direction of the neuron's preferred direction (summed direction vectors were applied if multiple neurons fired within one millisecond). At the end of each trial the cursor position was reset to the start location at $(0.25, 0.25)$.

\subheading{Details to: Hamiltonian dynamic improves network behavior at saddle points (Fig.~\ref{fig:saddle})}
A three-layer perceptron network consists of 784 input neurons, 30 hidden neurons and 10 output neurons are used to learn the MNIST data set. As shown in Fig.~\ref{fig:saddle}A, the activation function of the hidden layer and out layer are sigmoid function and Winner-Take-All (WTA) respectively. In each trial, one digital is chosen randomly from the MNIST data set as input. As the network gets immediate reward, no eligibility trace is used here. To be specific, the gradients of the expected reward $\ddthetai\log\mathcal{V}(\bth)$ for synapse $i$ are $r \, \hz_i \,  (z_i -f_i(u))$ if connected to neurons in the hidden layer and $r \, \hz_i \,  (z_i -g_i(u))$ otherwise, where $r$ is the binary reward, $\hz_i$ and $z_i$ denotes the outputs of presynaptic and postsynaptic neurons of synapse $i$. $u$ is the weighted sum of inputs to postsynaptic neuron of synapse $i$. In our simulation, we set the same learning rate 0.02 for both Hamiltonian sampling and synaptic sampling. The other parameters $a$, $c$ and $\beta$ is chosen to be 2, 2 and 0.2. In order to test whether
Hamiltonian dynamic can help to overcome saddle point problem, we first train the network with synaptic sampling and then continue to train it with Hamiltonian sampling (see Fig.~\ref{fig:saddle}B) with the current parameter setting. Note that the initial value of the momentum term is $0$. The result shows the network escaped from the current saddle point much faster with Hamiltonian dynamics.

\subheading{Details to: From reward-based learning to global network optimization (Fig.~\ref{fig:xor})}
A three-layer perceptron network consists of 784 input neurons, 30 hidden neurons and 10 output neurons are used to learn the MNIST data set. As shown in Fig.~\ref{fig:saddle}A, the activation function of the hidden layer and out layer are sigmoid function and soft Winner-Take-All (WTA) respectively. In each trial, one digit is chosen randomly from the MNIST data set as input. As the network gets immediate reward, no eligibility trace is used here. To be specific, the estimator of the gradients of the expected reward was directly given according to Eq.~\ref{eqn:eligibility-trace}, by
\begin{equation}
   \ddthetai \log \mathcal{V}(\bth) \;\approx\;  r(t) \, \hz_{\prei}(t) \,  (\zkt -f_{\posti}(t))
\end{equation}
In our simulation, we used the same learning rate 0.02 for both Hamiltonian sampling and synaptic sampling. The other parameters $a$, $c$ and $\beta$ were chosen to be 2, 2 and 0.2. In Fig~\ref{fig:xor}B we first trained the network with synaptic sampling for $10^6$ trials. We then continued training with Hamiltonian sampling using the parameter setting at that time point (initially the momentum term was $0$).

\section*{Acknowledgements}

The research leading to these results has received funding from the 
European Union Seventh Framework Programme (FP7/2007–2013) under EU 
grants $\#$604102 and $\#$720270 (Human Brain Project).

\bibliographystyle{unsrt}
\bibliography{548}

\end{document}